%% file: main.tex
\documentclass[10pt,twocolumn,letterpaper]{article}

\usepackage[pagenumbers]{cvpr} 
\usepackage{tabularx}
\usepackage{booktabs}
\usepackage{multirow}
\usepackage{graphicx}
\usepackage{siunitx}
\usepackage{xcolor}
\usepackage{stmaryrd}
\usepackage{makecell}
\usepackage{array}

\usepackage[utf8]{inputenc} 
\usepackage[T1]{fontenc}    
\usepackage[english]{babel} 

\usepackage{amsmath}
\usepackage{setspace}
\usepackage[ruled]{algorithm2e} 
\LinesNotNumbered 
\RestyleAlgo{ruled}
\SetStartEndCondition{ }{}{}%
\SetKwProg{Fn}{def}{\string:}{}
\SetKw{KwTo}{in}\SetKwFor{For}{for}{\string:}{}%
\SetKwIF{If}{ElseIf}{Else}{if}{:}{elif}{else:}{}%
\SetKwFor{While}{while}{:}{fintq}%
\AlgoDontDisplayBlockMarkers\SetAlgoNoEnd\SetAlgoNoLine%

\input{preamble}
\definecolor{cvprblue}{rgb}{0.21,0.49,0.74}
\usepackage[pagebackref,breaklinks,colorlinks,allcolors=cvprblue]{hyperref}

\def\paperID{5944} 
\def\confName{CVPR}
\def\confYear{2026}

\title{MeshRipple: Structured Autoregressive Generation of Artist-Meshes}

\author{
  {Junkai Lin}$^{1}$ \quad
  {Hang Long}$^{1}$ \quad
  {Huipeng Guo}$^{1}$ \quad
  {Jielei Zhang}$^{1}$ \quad
  {Jiayi Yang}$^{1}$ \quad
  {Tianle Guo}$^{1}$ \quad
  {Yang Yang}$^{1}$ \\
  {Jianwen Li}$^{2}$ \quad 
  {Wenxiao Zhang}$^{2}$ \quad
  {Matthias Nießner}$^{3}$ \quad
  {Wei Yang}$^{1, \dag}$
  \vspace{0.2cm}
  \\
  $^1$ Huazhong University of Science and Technology \quad 
  $^2$ Independent Researcher \\ 
  $^3$ Technical University of Munich \\
  \url{https://maymhappy.github.io/MeshRipple/}
}
\begin{document}
\twocolumn[{%
\maketitle
\begin{center}
    \vspace{-2mm}
    \centering
    \captionsetup{type=figure}
    \includegraphics[width=\textwidth]{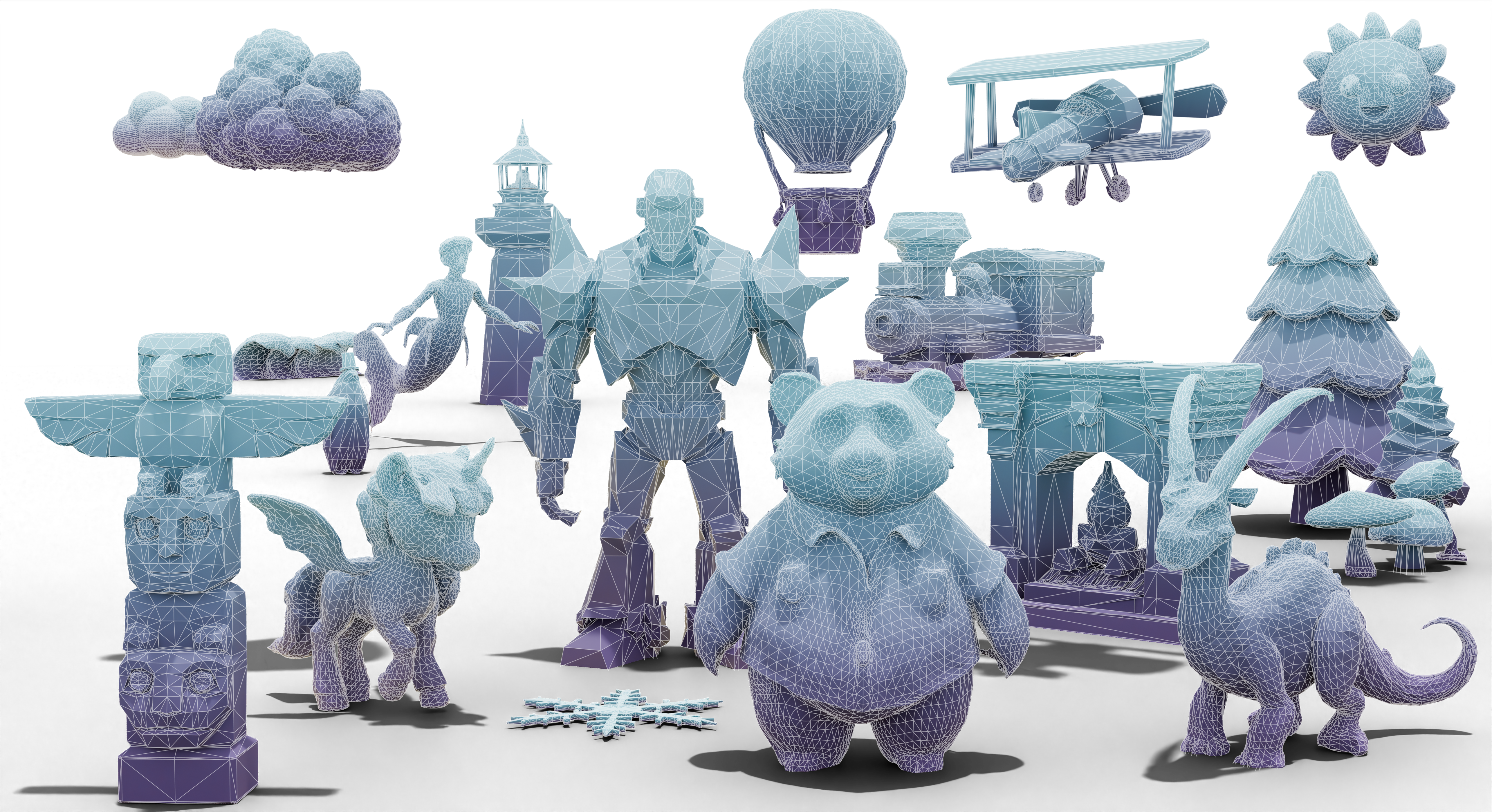}
    \captionof{figure}{\textbf{Gallery of results generated by MeshRipple.} Our method handles diverse mesh styles and topologies, producing artist-style meshes with optimized topology (airplane, train, totem) as well as dense smooth-surface meshes (bears, unicorn, dinosaur), all with coherent geometry and high-fidelity details.}
    \label{fig: teaser}
\end{center}%
}]

\renewcommand{\thefootnote}{\fnsymbol{footnote}}
\footnotetext{$\dag$ corresponding author: weiyangcs@hust.edu.cn}

\input{sec/0_abstract}
\input{sec/1_intro}
\input{sec/2_Related_Work}
\input{sec/3_method}

\input{sec/4_Experiments}

\input{sec/5_conclusion}
{
    \small
    \bibliographystyle{ieeenat_fullname}
    \bibliography{main}
}

\clearpage
\appendix
\setcounter{page}{1}
\maketitlesupplementary

\input{supplementary_material/1_details_of_tokenization_algorithm}

\input{supplementary_material/2_more_implementation_details}
\input{supplementary_material/3_more_ablation_study}
\input{supplementary_material/5_more_results}
\end{document}

%% file: sec/0_abstract.tex
\begin{abstract}
%
Meshes serve as a primary representation for 3D assets. Autoregressive mesh generators serialize faces into sequences and train on truncated segments with sliding-window inference to cope with memory limits. However, this mismatch breaks long-range geometric dependencies, producing holes and fragmented components. 
To address this critical limitation, we introduce \textbf{MeshRipple}, which expands a mesh outward from an active generation frontier, akin to a ripple on a surface.
MeshRipple rests on three key innovations: a frontier-aware BFS tokenization that aligns the generation order with surface topology; an expansive prediction strategy that maintains coherent, connected surface growth; and a sparse-attention global memory that provides an effectively unbounded receptive field to resolve long-range topological dependencies.
This integrated design enables MeshRipple to generate meshes with high surface fidelity and topological completeness, outperforming strong recent baselines.

\end{abstract}

%% file: sec/1_intro.tex
\section{Introduction}
Meshes are the native representation for high-quality 3D assets in film, game, and product design pipelines, and are supported by virtually all graphics tools and hardware. Directly generating editable, simulation-ready meshes can thus substantially reduce artist workload and accelerate industrial content creation. One approach is to generate continuous fields from latent codes with diffusion models and then extract surfaces via marching cubes~\cite{3dshape2vecset, clay, triposg, hunyuan3d, trellis, xcube, direct3d-s2, sparc3d}. While effective for geometry, this pipeline offers limited control over topology: marching cubes tends to produce overly dense, irregular, and non-artist-friendly meshes that require extensive post-processing.

Recent work instead employs autoregressive (AR) models to directly synthesize meshes with artist-like topology, often conditioned on input geometry such as point clouds. These methods tokenize meshes into sequences 
and learn a next-token distribution that respects mesh connectivity, leading to more structured and production-ready outputs. However, representing large, detailed meshes in sequence form inevitably yields excessively long token streams. To make AR training tractable, state-of-the-art approaches adopt a truncated-training strategy: the sequence is partitioned into fixed-length context windows; each window is trained independently; and, at inference, a rolling KV cache is used to approximate longer-range context~\cite{zhao2025deepmesh, liu2025quadgptnativequadrilateralmesh, liu2025meshrftenhancingmeshgeneration}. Yet existing tokenization schemes, including coordinate-based orderings~\cite{nash2020polygen,siddiqui2024meshgpt,chen2024meshxl,liu2025quadgptnativequadrilateralmesh}, depth-first expansion~\cite{tang2024edgerunner, lionar2025treemeshgpt}, and block-/patch-based encodings~\cite{tang2024edgerunner,DBLP:BPT, zhao2025deepmesh}, do not guarantee that the relevant elements fall within the truncated window. As a result, the model is frequently forced to predict connectivity without seeing the correct local neighborhood, leading to broken surfaces and fragmented components. Rolling the KV cache at inference time cannot fundamentally fix this issue, since the model was never consistently trained with the true context.
Recent advances, such as DeepMesh~\cite{zhao2025deepmesh}, MeshRFT~\cite{liu2025meshrftenhancingmeshgeneration}, and QuadGPT~\cite{liu2025quadgptnativequadrilateralmesh}, improve performance via preference-based fine-tuning with reinforcement learning. While these approaches enhance topological quality and style alignment, they do not fully resolve connectivity errors.

In this work, we introduce \textbf{MeshRipple}, a topology-aligned autoregressive framework for generating large, artist-friendly triangle meshes with high structural fidelity. The core component is Ripple Tokenization (RT), which couples a breadth-first face ordering with an explicit, dynamically maintained frontier:
we traverse faces in half-edge based BFS order;
at each step we maintain the set of faces that bound at least one unvisited neighbor (the frontier); and
for every newly added face we record a root pointer to the frontier face from which it grows.
By construction, the structurally relevant tokens for the next face prediction occupies the most recent portion of the sequence. Hence, a truncated window that covers the tail segment containing the frontier reliably includes the local context required for the next face, directly mitigating the topology–context mismatch that limits prior tokenizations.

Building on RT, we design an autoregressive transformer tailored for structured mesh growth. A compact hourglass backbone first compresses fine-grained vertex tokens into per-face embeddings. Upon which, at each step, the model jointly predicts the the grown face attached to the current root, and the next root to expand, constraining generation to expand a coherent surface rather than freely sampling incompatible triangles. To incorporate longer-range cues (e.g., symmetry and repetition) without prohibitive memory, we introduce Native Sparse Contextual Attention (NSCA): a causal, sparsified cross attention mechanism that queries a compressed summary of the full mesh history and selectively retrieves detailed relevant tokens. NSCA complements the local truncated window while keeping computation and memory bounded.
Through this structure-aware design, MeshRipple aligns tokenization, truncated training, and autoregressive inference. The result is a substantial reduction in broken surfaces and artifacts, enabling high-face-count mesh generation at scale (see~\Cref{fig: teaser}). 

\noindent In summary, our contributions are three-fold:
\begin{itemize}
    \item \textbf{Ripple Tokenization.} A topology-aligned face traversal with an explicit frontier that concentrates the relevant tokens into the tail of the sequence, aligning tokenization with truncated autoregressive training.
    \item \textbf{Expansive Prediction Strategy.} A decoding scheme that jointly predicts the root and the attached face, enforcing face frontier connectivity.
    \item \textbf{Native Sparse Contextual Attention (NSCA).} A causal sparse attention module that retrieves long-range mesh cues, supporting scale-up generation of meshes.
\end{itemize}

%% file: sec/2_Related_Work.tex
\begin{figure*}[t]
\centering
\includegraphics[width=\textwidth]{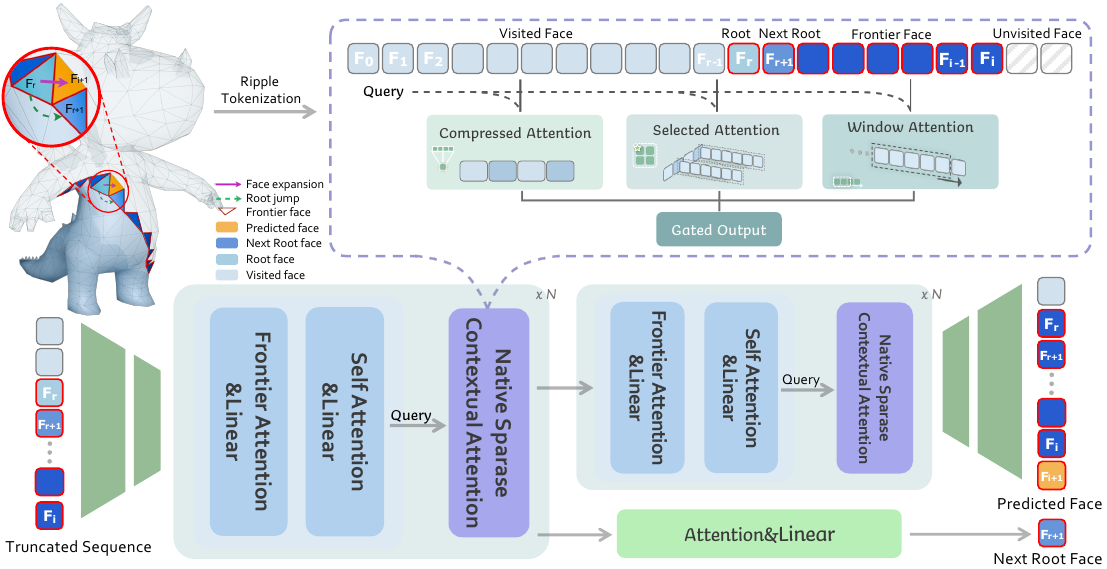}
\caption{\textbf{Overview of MeshRipple.} The input mesh is first serialized into a token sequence via \textbf{Ripple Tokenization}, which is then truncated into fixed-length segments as input to a structured autoregressive model. The model employs hourglass layers at both ends to convert between vertex and face tokens. The core consists of a stack of $2\times N$ identical blocks, each comprising a \textbf{Frontier-Attention} layer, a self-attention layer, a cross-attention layer for point-cloud conditioning (omitted for clarity), and a \textbf{Native Sparse Contextual Attention} layer that attends to the full mesh sequence under a causal mask. The middle hidden states are additionally fed into a lightweight head to predict the next root face to expand.
}
\vspace{-1em}
\label{fig: pipeline}
\end{figure*}

\section{Related Work}
\label{sec:related}

\paragraph{3D Shape Generation.}
Early work on 3D generation, constrained by scarce 3D data, largely follows two strategies. One line adopts SDS-based optimization from image diffusion models~\cite{poole2022dreamfusion, DBLP:Fantasia3D, DBLP:Text-to-3D, li2023sweetdreamer, DBLP:Magic3D, DBLP:ScaleDreamer, DBLP:DreamBooth3D, DBLP:DreamCraft3D, DBLP:DreamGaussian, DBLP:ScoreJacobianChaining, DBLP:AnimatableDreamer, wang2023prolificdreamerhighfidelitydiversetextto3d, DBLP:GaussianDreamer}, while another predicts multi-view images and then reconstructs 3D geometry from them~\cite{chen2024v3d, DBLP:Animate3D, DBLP:One-2-3-45, liu2023zero1to3zeroshotimage3d, DBLP:SyncDreamer, DBLP:Wonder3D, shi2023zero123++, DBLP:SV3D, DBLP:Consistent123, DBLP:Unique3D, DBLP:Hi3D, zhao2024flexidreamer, shi2024mvdreammultiviewdiffusion3d, Richdreamer, wang2023imagedream, DBLP:DreamReward}.
LRM~\cite{Lrm} then introduces a transformer-based reconstructor that predicts a NeRF representation~\cite{DBLP:NeRF} from a single image, and subsequent work~\cite{Instant3d, DBLP:Meta3DAssetGen, DBLP:LGM, DBLP:PF-LRM, DBLP:CRM, DBLP:DMV3D, DBLP:GRM, DBLP:GeoLRM, DBLP:GS-LRM, ziwen2025longlrm, DBLP:TriplaneMeetsGaussianSplatting} improves this paradigm with multi-view supervision and alternative 3D parameterizations such as 3D Gaussians~\cite{DBLP:3dgs}. 
In parallel, early 3D-native diffusion models~\cite{gao2022get3dgenerativemodelhigh, jun2023shapegeneratingconditional3d, DBLP:MeshDiffusion, nichol2022pointegenerating3dpoint, chang20243d} operate directly on point clouds, but are limited by dataset scale and generalization. 
3DShape2VecSet~\cite{3dshape2vecset} instead encodes point clouds into a latent set and learns a generative distribution in this space, and CLAY~\cite{clay} demonstrates that this paradigm can be scaled up effectively. Subsequent methods~\cite{DBLP:3DTopiaxl, li2024craftsman3d, DBLP:Direct3D, triposg, hunyuan3d, ye2025hi3dgen} further enrich the latent representation with textures and normals, or refine geometric detail via normal supervision.
Trellis~\cite{trellis} adopts a more structured latent space based on sparse voxels, while Direct3D-S2~\cite{trellis} and Sparc3D~\cite{sparc3d} extend this idea using multi-resolution latents or higher-resolution voxel decoders.
%
%
%
The aforementioned methods generate 3D assets in implicit or volumetric form and only obtain meshes via post-hoc extraction (e.g., Marching Cubes~\cite{marchingcube}). The resulting faces are often overly dense and require significant post-precessing.

\noindent \textbf{Mesh Tokenization and Autoregressive Generation.}
Autoregressive models that generate meshes directly have attracted growing interest. PolyGen~\cite{nash2020polygen} predicts vertices and edges separately, while MeshGPT~\cite{siddiqui2024meshgpt} pioneers the combination of a VQ-VAE~\cite{DBLP:vqvae} with an autoregressive transformer for mesh synthesis; later works~\cite{chen2024meshxl, chen2024meshanything, DBLP:LLaMA-Mesh, DBLP:PivotMesh} explore alternative architectures and conditional settings. In parallel, the community has rapidly shifted toward explicit mesh quantization and tokenization: MeshAnythingv2~\cite{chen2025meshanything} introduces Adjacent Mesh Tokenization (AMT), EdgeRunner~\cite{tang2024edgerunner} derives a sequence via an EdgeBreaker-style traversal~\cite{rossignac2002edgebreaker}, BPT~\cite{DBLP:BPT} adopts a patchified, blockwise strategy, and Mesh Silksong~\cite{song2025mesh} tokenizes edges with a BFS-style traversal. These tokenizers primarily aim to maximize compression to alleviate memory constraints. MeshTron~\cite{hao2024meshtron} instead operates directly on vertex coordinates and employs an hourglass architecture for compression. DeepMesh~\cite{zhao2025deepmesh} and MeshRFT~\cite{liu2025meshrftenhancingmeshgeneration} introduce reinforced fine-tuning to improve topology quality, and QuadGPT~\cite{liu2025quadgptnativequadrilateralmesh} extends to quadrilateral meshes. Despite this progress, these approaches still suffer from broken surfaces. 

%% file: sec/3_method.tex
\vspace{-0.2em}
\section{Method}

Our framework aims to generate large, structurally coherent meshes under truncated autoregressive (AR) training. It consists of two key components: (i) \textbf{Ripple Tokenization}, which couples a breadth-first face ordering with an explicit dynamically maintained frontier, such that the structurally relevant context for the next face prediction is concentrated near the tail of the sequence; and (ii) \textbf{an structured autoregressive transformer} with frontier attention and a lightweight context attention that jointly predicts the next face attached to current root face, and the subsequent root face on the frontier to expand.
\subsection{Ripple Tokenization}
We conduct a half-edge based breadth-first traversal of faces with an explicit, dynamically maintained frontier queue. At each step, we grow new faces from a set of frontier faces that bound the already generated region. This construction ensures that, for any next face to be predict, its incident neighborhood lies within the most recent portion of the sequence, making truncated training well-aligned with mesh topology.
Given a triangle mesh $\mathcal{M} = (\mathcal{V}, \mathcal{F})$, we follow standard practice as in~\cite{hao2024meshtron,zhao2025deepmesh} and discretize vertex coordinates, by normalizing $\mathcal{V}$ into a canonical bounding box and uniformly quantize each coordinate. Each face $\mathbf{F}_i \in \mathcal{F}$ is represented by three quantized vertices, $F_i = [\mathbf{v}^i_0, \mathbf{v}^i_1, \mathbf{v}^i_2]$.
To support consistent traversal and ensure the consistency of generated face normals, we orient each face in a counterclockwise order with respect to its outward normal, and derive its three directed half-edges:
$(\mathbf{v}^i_0 \!\rightarrow\! \mathbf{v}^i_1),\;
(\mathbf{v}^i_1 \!\rightarrow\! \mathbf{v}^i_2),\;
(\mathbf{v}^i_2 \!\rightarrow\! \mathbf{v}^i_0).$
Therefore, two faces are considered adjacent if they share a pair of opposite half-edges (a half-edge and its twin).
\begin{figure}[t] 
\centering 
\includegraphics[width=\columnwidth]{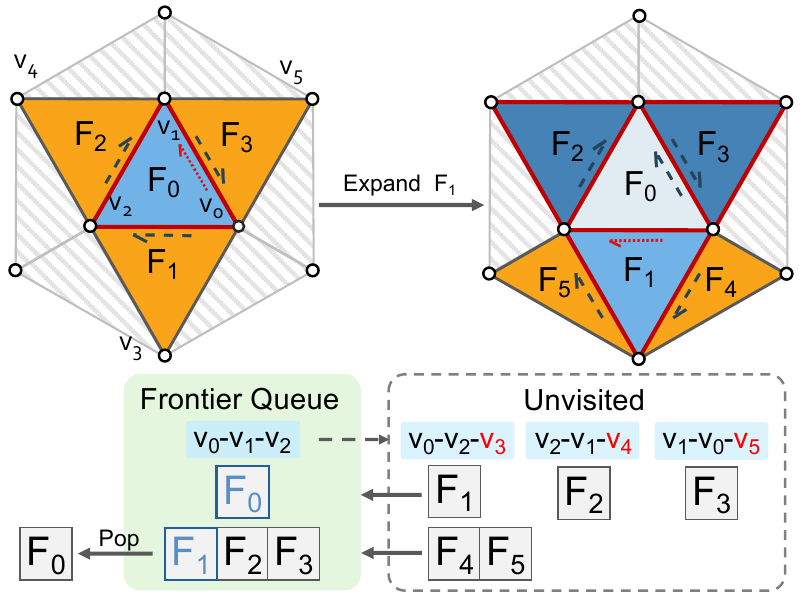} 
\caption{\textbf{Illustration of Ripple Tokenization.} At each step, the current root face expands following the counterclockwise half-edge order. Faces that still have unvisited neighbors remain in the FIFO frontier queue; once all neighbors are visited, the face is popped from the queue.}
\label{fig:rp}
\vspace{-1.5em}
\end{figure}
\subsubsection{BFS Traversal}

Starting from an initial seed face $\mathbf{F}_s$, we perform a breadth-first ordering over faces using the half-edge connectivity.
For a face, we look at its three directed half-edges in their fixed (e.g., counterclockwise) order. For each half-edge, if there is an adjacent face across its twin that has not been visited, we mark this neighbor as visited, append it to the output sequence $\mathcal{S}$, and schedule it to be expanded later. In this way, new faces are discovered and inserted into the sequence following the local half-edge order of their root face, giving a deterministic, topology-respecting BFS ordering.
%
%

\noindent \textbf{Non-manifoldness.}
Our method inherently handles non-manifold edges, i.e., when an edge is incident to more than two faces, our half-edge structure records all corresponding twins and each incident face is simply treated as a valid neighbor. For non-manifold vertices, our algorithm naturally divides them into separate disconnected components.
%

\noindent \textbf{Disconnected Components.}
For meshes with multiple connected components, we process each component independently with the BFS procedure.
Between components, we insert a special control token \colorbox{GreenYellow}{\textbf{N}} into $\mathcal{S}$.

\subsubsection{Frontier Queue and Root-Face Registration}

To make the traversal explicitly compatible with truncated AR training, we maintain a FIFO frontier queue $\mathcal{B}_i$ for each face $\mathbf{F}_i$. 
Concretely during the BFS traversal, when a face $\mathbf{F}_i$ is appended to $\mathcal{S}$, we record a root index $r_i$: $r_i \in \{1, \dots, |\mathcal{S}|\}$, indicating from which face $\mathbf{F}_{r_i} \in {\mathcal{B}}$ the new face $\mathbf{F}_i$ is grown (i.e., which shared edge induces its attachment). And if all neighbors of $\mathbf{F}_{r_i}$ have been visited, we remove it from $\mathcal{B}_i$. Then we add $\mathbf{F}_i$ into $\mathcal{B}_{i}$, it forms a new frontier $\mathcal{B}_{i+1}$ for $\mathbf{F}_{i+1}$.  
%
Hence, for each face $\mathbf{F}_i$ in mesh sequence $\mathcal{S}$, we have recorded:
\begin{equation}
    \mathbf{F}_i \mapsfrom  \{ r_i, \mathcal{B}_i \}
\end{equation}
Notice that $\mathbf{F}_{r_i}$ is always the first face in $\mathcal{B}_i$ due to BFS. This design ensures that the relevant context (the active frontier) is always located near the end of the sequence. And a truncated window that covers the last $L$ tokens thus contains the necessary context to predict the next face (see~\Cref{fig:rp} for detailed explanation). Also,
the root index $r_i$ restricts autoregressive sampling to expand from valid boundary locations, substantially reducing topological ambiguities and broken surfaces. Existing tokenization schemes either follow a coordinate-ordered traversal~\cite{siddiqui2024meshgpt,chen2024meshxl,hao2024meshtron}, depth-frist expansion~\cite{tang2024edgerunner, lionar2025treemeshgpt} or a patch-based encoding~\cite{DBLP:BPT,zhao2025deepmesh}, important tokens can easily fall outside the truncated input sequence. The most related work to ours is the concurrent work Mesh Silksong~\cite{song2025mesh}, which also adopts a breadth-first traversal. However, we differ fundamentally as Mesh Silksong tokenize both vertices and edges, resulting in an excessively large vocabulary, and requires preprocessing for non-manifoldness. In contrast, our method is more flexible: it explicitly maintains a frontier set of boundary faces, simplifies the prediction space, and naturally extends to non-manifold configurations.

\subsection{Structured Autoregressive Model}
Given a mesh encoded by Ripple Tokenization, our goal is to autoregressively generate faces in a way that respects the frontier-based growth process while compatible with truncated training. To this end, we design a structured autoregressive transformer equipped with three key components: a \textbf{Frontier Attention} mechanism that emphasizes structurally relevant frontier faces, a \textbf{Native Sparse Contextual Attention} (NSCA) module that injects long-range mesh context under strict memory constraints, and an \textbf{Expansive Prediction} scheme that jointly predicts the new face grown from the current root and the next root face to expand. 
Please see~\Cref{fig: pipeline} for the detailed framework.
Starting from the quantized coordinate sequence $\mathcal{S}$, we adopt an outer hourglass-style transformer to progressively merge: coordinate-level tokens into vertex-level embeddings, and finally into face-level tokens, where each token corresponds to one face. Subsequent conditioning, frontier and contextual attentions are applied at the face-token level.

%


\noindent \textbf{Truncated Training}
High-poly meshes induce long face sequences. We follow the standard truncated training regime as in MeshTron~\cite{hao2024meshtron} and DeepMesh~\cite{zhao2025deepmesh} and partition the mesh sequence into fixed-size windows, with padding applied to insufficient-length segments. 
\subsubsection{Frontier Attention}
For each face $\mathbf{F}_i$, let $\mathcal{B}_i$ denote the FIFO frontier queue constructed during tokenization. Faces in $\mathcal{B}_i$ are precisely those from which valid growth can occur.
To emphasize this structure, we introduce a frontier attention layer. It uses standard multi-head attention over the windowed input tokens, but with a mask that restricts attention toward the current frontier:
for the query at position $i$, only tokens corresponding to faces in $\mathcal{B}_i$ (and not beyond $i$) are assigned non-negative logits; all others are masked out. 
\begin{equation}
\mathbf{M}^{\mathrm{b}}_{i,j} =
\begin{cases}
0,  & \mathbf{F}_j \in \mathcal{B}_i \text{ and } j \le i \\
-\mathrm{Inf}, & \text{otherwise}
\end{cases}
\end{equation}
This encourages the model to primarily from the active growth boundary when predicting the next face, and route gradients to the subset of tokens that are structurally responsible for connectivity.
%
%
%



\subsubsection{Native Sparse Contextual Attention}

While frontier attention provides accurate local structure, many meshes exhibit long-range patterns.
Conventional sliding-window inference approximates this with KV rolling caches suffers training-inference inconsistency problem. To address, we propose NSCA, which offers causal, memory-efficient access to the full mesh history. 

For the given full mesh sequence $\mathcal{S}$, we first obtain coordinate-level embeddings, aggregate them into vertex-level embeddings, and finally to a per-face token by catenating along the channel dimension, which is finally projected by a small MLP to a pre-defined hidden dimension. 
We partition the full face sequence into non-overlapping blocks.
Note that the query tokens in the main model are taken from the truncated window, whereas the keys and values for NSCA are computed from the full sequence, so their lengths differ. To ensure causal consistency, we construct masks at both the block and token levels: for any block that contains tokens positioned after the current step, we assign its entire block in the mask with $-\text{inf}$, so it cannot be attended to when predicting the current token.
Following the original DeepSeek's Native Sparse Attention~\cite{yuan2025nativesparseattention}, each valid block is mapped to a single compressed key–value pair. Using the hidden state of the main autoregressive model as query, we compute an importance score for each compressed block token and select the top-$k$ blocks, for which we then retrieve the original (uncompressed) keys and values. In parallel, we maintain a standard local window of each token. A block-level causal mask is applied to all compressed and selected tokens, while a token-level causal mask is used for the local window. Finally, a lightweight router learns to weight and fuse three sources—the compressed block tokens, the detailed tokens from the top-$k$ selected blocks, and the local window—and the resulting mixture is used as the contextual memory attended by the query.

\subsubsection{Expansive Prediction}

Unlike prior autoregressive models that only predict the next face token, we adopt an \emph{expansive prediction} strategy: at each step, the model jointly predicts \textbf{the new face} grown from the current root, and \textbf{the next root face} to expand on the frontier.
Concretely, we attach an additional root prediction head to middle of the network. 
Instead of predicting the next root face token directly, we let the model predict how many positions to move along the FIFO frontier from the current root to obtain the next root.
%
Formally, let $\mathbf{F}_i$ and $\mathbf{F}_{i+1}$ are two adjacent faces in input sequence, and let $r_i$ and $r_{i+1}$ be their root face indices respectively. We define the offset steps as $\Delta_{i + 1} = r_{i+1} - r_i$. Notice $\Delta_{i + 1}=0$ if $\mathbf{F}_i$ and $\mathbf{F}_{i+1}$ share the same root.

\noindent \textbf{Training.} A lightweight pointer head attends over the middle layer hidden states to produce $p_\theta(\Delta_{i + 1})$, and the root loss is a standard cross-entropy
\begin{equation}
    \mathcal{L}_{\text{root}} \;=\; -\sum_{i} \log p_\theta(\Delta_{i + 1} | \mathbf{F}_t, t \leqslant	i).
\end{equation}
The overall training objective combines next root prediction loss and the standard next face prediction loss.

\noindent \textbf{Inference.} For inference, at each step we maintain a dynamic FIFO frontier queue $\mathcal{B}$, the current generated mesh sequence, and the current root face $\mathbf{F}_r$.
To predict the next face, we enforce structural consistency by masking out all candidate faces that do not attach to $\mathbf{F}_r$, and then choose the final prediction $\mathbf{F}^\ast$ from the remaining candidates using sampling-based decoding. We append $\mathbf{F}^\ast$ to the full mesh sequence (used by NSCA) and to the autoregressive input, and update the frontier queue by adding $\mathbf{F}^\ast$.
Meanwhile, the root pointer head predicts $\Delta$, which selects the next root face $\mathbf{F}_r^\prime$ by moving forward $\Delta$ steps on the frontier. 
For the \colorbox{GreenYellow}{\textbf{N}} token, we simply lift the connectivity constraint.
We then update the root face $\mathbf{F}_r$ to $\mathbf{F}_r^\prime$, pop out all faces before $\mathbf{F}_r^\prime$ from $\mathcal{B}$. 
The frontier attention mask and causal attention mask for NCSA are updated accordingly, and conduct the next round prediction.

\begin{figure*}[t]
\centering
\includegraphics[width=\textwidth]{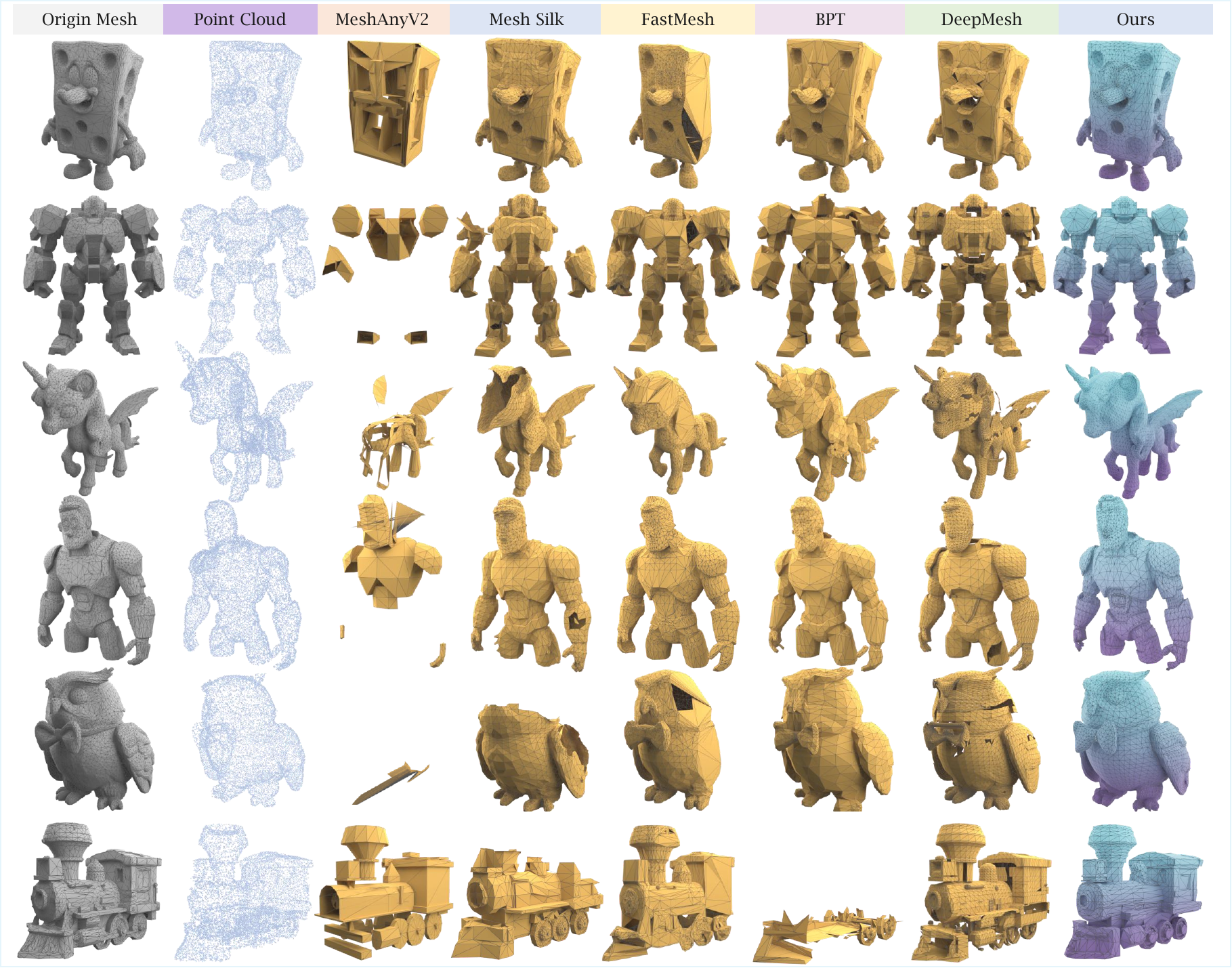}
\caption{\textbf{Qualitative comparison of point cloud-conditioned generation between MeshRipple and baselines.} Baselines inevitably produce broken surfaces and holes, whereas MeshRipple yields more complete and coherent geometries.} 
\vspace{-1em}
\label{fig: compare}
\end{figure*}

%% file: sec/4_Experiments.tex
\begin{table}[t]
    \centering
    \small 
    \setlength{\tabcolsep}{0pt} 

    \caption{
        \textbf{Quantitative comparison on Dense-Mesh and Artist-Mesh datasets.} 
        We scale CD by $10^3$. 
        The best result is highlighted in \textbf{bold}, and the second-best is \underline{underlined}.
    }
    \label{tab:compare_quantitative_results}

    \begin{tabular*}{\columnwidth}{@{\extracolsep{\fill}} l l c c r }
        \toprule
        Dataset & Method & CD($\downarrow$) & HD($\downarrow$) & NC($\uparrow$) \\
        \midrule
        
        \multirow{6}{*}{\shortstack[l]{Dense\\Mesh}} 
          & MeshAnythingV2 & 109.64 & 0.2314 & -0.0096 \\
          & BPT            & 60.19  & 0.1089 & \underline{0.6066} \\
          & FastMesh       & 64.62  & 0.1155 & 0.0002 \\
          & Mesh SilkSong  & 61.99  & 0.1435 & 0.5529 \\
          & DeepMesh       & \underline{50.27} & \textbf{0.0893} & 0.6025 \\
          & Ours           & \textbf{48.73} & \underline{0.1057} & \textbf{0.6280} \\
        
        \midrule
        
        \multirow{6}{*}{\shortstack[l]{Artist\\Mesh}} 
          & MeshAnythingV2 & 68.11  & 0.1433 & -0.0504 \\
          & BPT            & 54.78  & 0.1014 & \underline{0.5084} \\
          & FastMesh       & \underline{47.26} & \underline{0.0972} & 0.0106 \\
          & Mesh SilkSong  & 49.72  & 0.1019 & 0.4709 \\
          & DeepMesh       & 51.11  & 0.1023 & 0.3174 \\
          & Ours           & \textbf{46.68} & \textbf{0.0938} & \textbf{0.5166} \\ 
        
        \bottomrule
    \end{tabular*}
\end{table}
\section{Experiments}

\noindent \textbf{Datasets.}
We construct our dataset from Objaverse-XL~\cite{deitke2023objaverse-xl}, G-Objaverse~\cite{zuo2024sparse3d}, ShapeNet~\cite{chang2015shapenetinformationrich3dmodel}, Toys4K~\cite{stojanov2021using}, and 3D-FUTURE~\cite{3d-future}. We discard meshes with more than 20k or fewer than 500 faces, excessive discretization artifacts (over-aggressive vertex merging), more than 20 connected components, or an intersection ratio above 10, yielding roughly 300k meshes. For artist-mesh evaluation, we hold out 300 meshes from G-Objaverse and 50 each from Toys4K and ShapeNet. For dense-mesh evaluation, we generate 500 meshes using Trellis~\cite{trellis}.
%

\noindent \textbf{Implementation Details.}
The architecture follows a \([4,4,9]\) Hourglass configuration with an embedding dimension of 1024, a 100-class next-root prediction head, and vertex quantization into 256 bins. For context injection, we use a 256-dimensional embedding that is concatenated along the channel dimension. For each mesh, we sample 40,960 points and randomly select 16,384 points as the condition. These point clouds are encoded into latent features by a Michelangelo encoder, which are then injected via cross-attention. We set the maximum mesh length to 20k faces, and the input sequence 1k faces. We train our model with AdamW optimizer, 
using a cosine learning rate schedule decays from \(1\times10^{-4}\) to \(1\times10^{-5}\) with weight decay \(0.1\). Training takes 16 days on 16 NVIDIA A800 GPUs.

\begin{figure}[t]
    \centering
    \includegraphics[width=\linewidth]{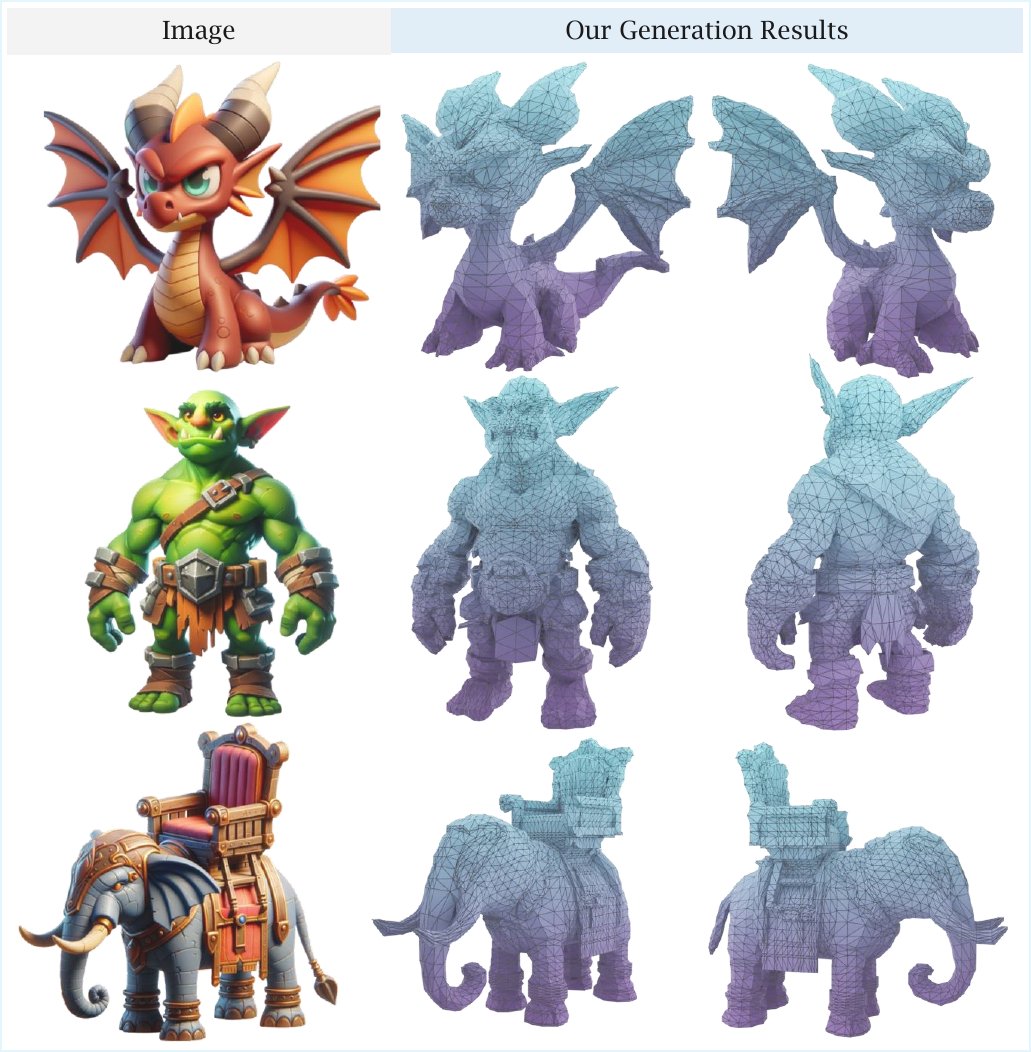}
    \caption{\textbf{Image-conditioned generation results}. Our method can generate high-fidelity meshes aligned with the input images.}
    \label{fig:image_condtion}
    \vspace{-1em}
\end{figure}

\subsection{Quantitative Analysis}
We quantitatively compare our method with MeshAnythingv2~\cite{chen2025meshanything}, BPT~\cite{DBLP:BPT}, FastMesh~\cite{kim2025fastmesh}, Mesh Silksong~\cite{song2025mesh}, and DeepMesh~\cite{zhao2025deepmesh} on two benchmarks: artist meshes and dense meshes. For fair evaluation, we uniformly sample 1{,}024 points from each generated mesh and its ground-truth counterpart, and report Chamfer Distance (CD), Hausdorff Distance (HD), and Normal Consistency (NC). As summarized in \Cref{tab:compare_quantitative_results}, our method achieves the best overall scores on the artist-mesh benchmark and performs competitively with DeepMesh on dense meshes, despite substantially lower compute: DeepMesh is pre-trained on 128 NVIDIA A100 GPUs for 4 days plus DPO post-training, whereas our model is trained on 16 A800 GPUs for 16 days.
\subsection{Qualitative Analysis}
\paragraph{Point-cloud Conditioned.}
We evaluate point cloud conditioned generation against five recent open-source baselines: MeshAnythingv2, BPT, FastMesh, Mesh Silksong, and DeepMesh. As shown in \Cref{fig: compare}, competing methods often miss fine topological structures and exhibit holes or fragmented geometry, whereas our meshes preserve both detail and global coherence. Our results are simultaneously more intricate than those of MeshAnything v2, BPT, FastMesh, and Mesh Silksong, and more structurally stable than those of DeepMesh. We attribute these gains to our topology-aware architecture, which couples global context modeling and point-cloud conditioning with an improved BFS expansion scheme based on a frontier mask.
\begin{figure}[t]
    \centering
    \includegraphics[width=\linewidth]{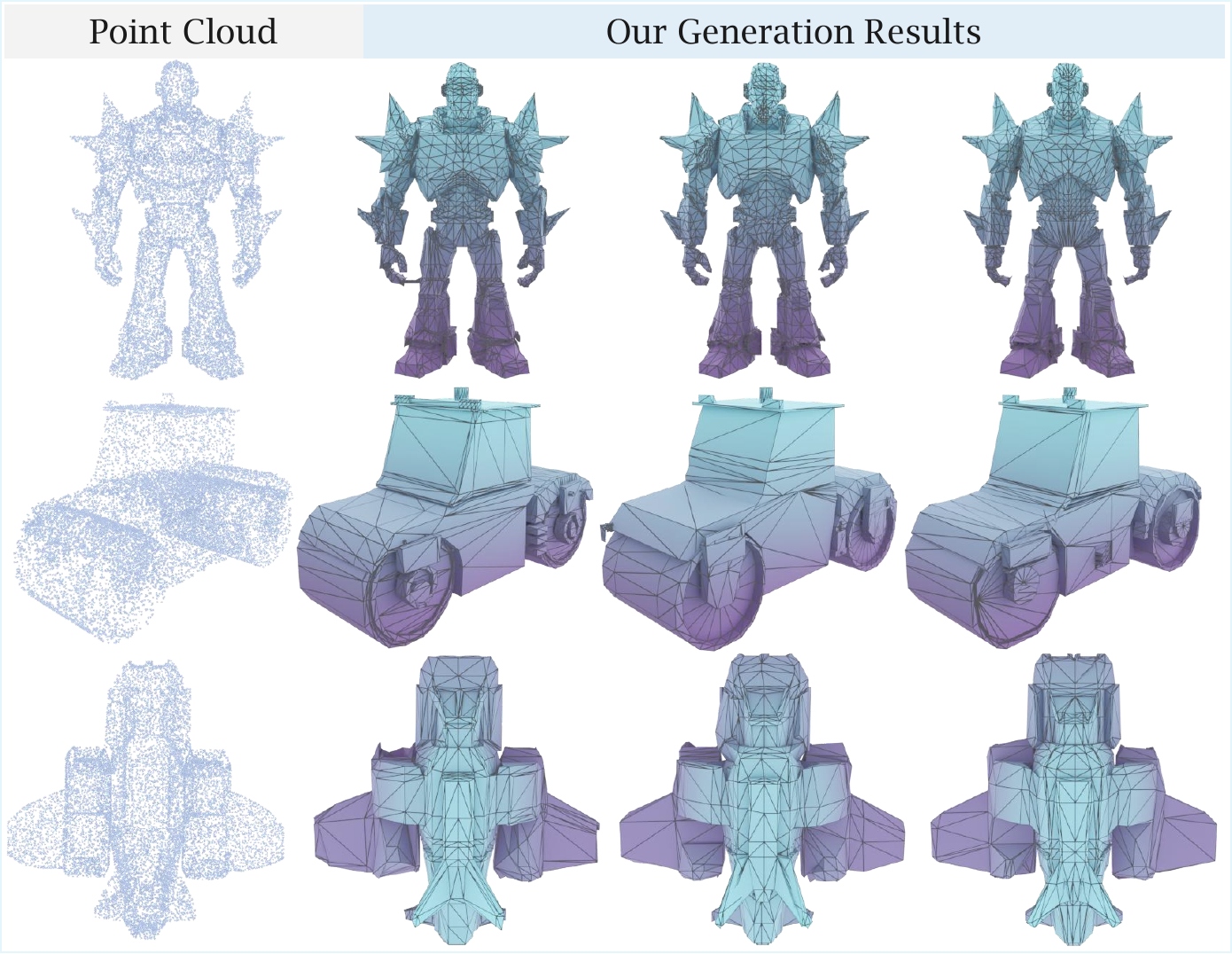}
    \caption{\textbf{Diversity of generations.} MeshRipple can generate meshes with diverse topologies given the same point cloud.}
    \vspace{-12pt}
    \label{fig:diversity}
\end{figure}

\noindent \textbf{Image Conditioned.}
For image conditioned generation, we adopt a two-stage pipeline. We first use TRELLIS~\cite{trellis} to synthesize a 3D mesh from the input image, then extract a point cloud from this mesh to condition our point cloud–based generator. Representative high-quality results are shown in~\Cref{fig:image_condtion}.


\begin{figure*}[t]
\centering
\includegraphics[width=\textwidth]{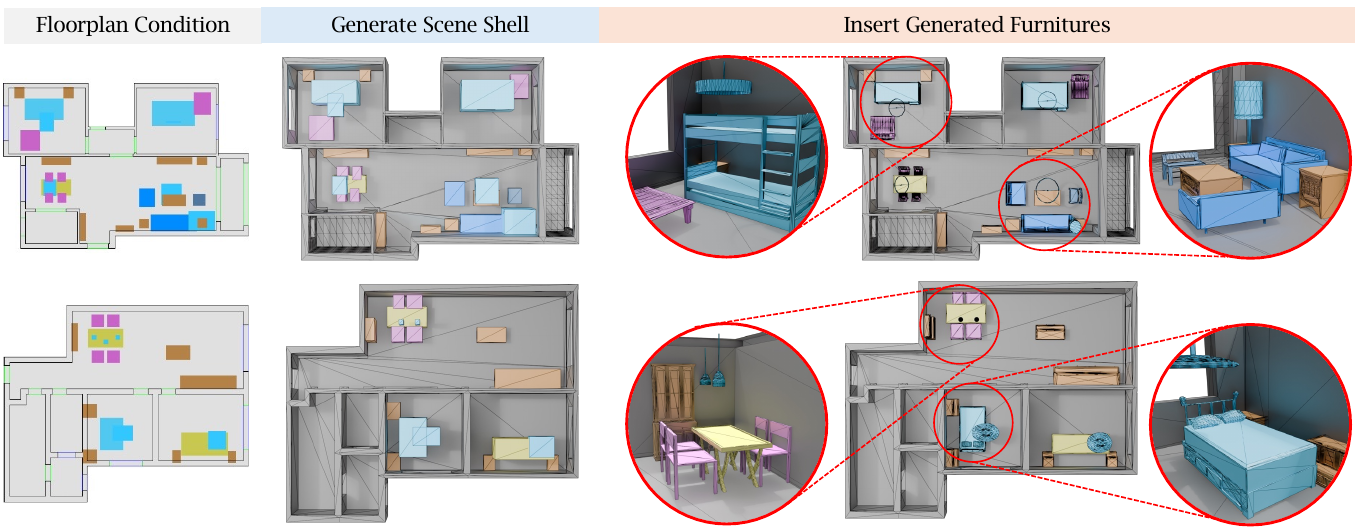}
\caption{\textbf{Scene generation results.} MeshRipple supports topology variations, non-manifold geometres, and multi-component assemblies. We demonstrate this through progressive scene generation. Trained on 3D-FRONT, we first generate a room shell from a floorplan image, and predict category-labeled bounding-box placeholders for furniture; we then generate and insert each furniture mesh independently.
}
\vspace{-16pt}
\label{fig: compare_room}
\end{figure*}

\noindent \textbf{Diversity.}
We assess the stochasticity of our generator by sampling multiple meshes from the same point-cloud input. As shown in~\Cref{fig:diversity}, the model produces a diverse set of plausible outputs while consistently preserving the underlying structure, demonstrating its ability to achieve both variation and high-fidelity synthesis.
\begin{table}[h]
  \centering
  \small 

  \caption{\textbf{Ablations of our design choices}. Best results shown in \textbf{bold}, second-best is \underline{underlined}.
  }
  \label{tab:ablation_subtractive}

  \begin{tabular*}{\linewidth}{l @{\extracolsep{\fill}} cc cc}
    \toprule
    \multirow{2}{*}{Method} & \multicolumn{2}{c}{CD} & \multicolumn{2}{c}{HD} \\
    \cmidrule(lr){2-3} \cmidrule(lr){4-5}
    & Value $\downarrow$ & $\Delta$ & Value $\downarrow$ & $\Delta$ \\
    \midrule
    
    Full Model (Ours) & \underline{52.67} & - & \underline{0.1058} & - \\
    \midrule
    
    w/o Frontier Mask & 54.79 & +2.12 & 0.1199 & +0.0141 \\
    w/o Context Injection & 60.56 & +7.89 & 0.1176 & +0.0118 \\
    w/o Non-manifold & 57.75 & +5.08 & 0.1193 & +0.0135 \\
    w/o NSCA & \textbf{52.43} & -0.24 & \textbf{0.1025} & -0.0033 \\
    w/o Root Constraint & 65.08 & +12.41 & 0.1180 & +0.0122 \\
    
    \bottomrule
  \end{tabular*}
  \vspace{-1.5em}
\end{table}

\definecolor{cabinet_furniture_color}{RGB}{183, 127, 66}
\definecolor{bed_furniture_color}{RGB}{85, 196, 240}
\definecolor{chair_furniture_color}{RGB}{200, 100, 200}
\definecolor{table_furniture_color}{RGB}{184, 200, 102}
\definecolor{sofa_furniture_color}{RGB}{0, 140, 255}
\definecolor{stool_furniture_color}{RGB}{80, 119, 160}
\definecolor{lamp_furniture_color}{RGB}{49, 198, 254}
\noindent \textbf{Scene Generation on 3D-FRONT.}
Our method supports diverse frontier configurations arising from topology variation, non-manifold geometry, and multi-component assemblies.%
To demonstrate, we train on 3D-FRONT~\cite{huan20253dfront}, which features varied topologies and component configurations. We replace each furniture instance with a category-labeled bounding-box placeholder to enable progressive prediction.
We replace the component segment control token \colorbox{GreenYellow}{\textbf{N}} with semantic tokens \colorbox{cabinet_furniture_color}{\texttt{CABINET}}, \colorbox{bed_furniture_color}{\texttt{BED}}, \colorbox{chair_furniture_color}{\texttt{CHAIR}}, \colorbox{table_furniture_color}{\texttt{TABLE}},  \colorbox{sofa_furniture_color}{\texttt{SOFA}}, \colorbox{stool_furniture_color}{\texttt{STOOL}}, and \colorbox{lamp_furniture_color}{\texttt{LAMP}}. 
At test time, we first generate a room shell from a floorplan and predict category-labeled bounding-box placeholders; we then generate each furniture mesh independently with MeshRipple, finetuned with ShapeNet, and place it into the corresponding location. High-quality results are shown in \Cref{fig: compare_room}.

\begin{table}[h!]
    \centering
    \small 
    \setlength{\tabcolsep}{0pt} 
    
    \caption{\textbf{Ablation on root prediction strategies.} We compare predicting absolute coordinates versus index offset.}
    \label{tab:ablation_root_pred}
    
    \begin{tabular*}{\linewidth}{@{\extracolsep{\fill}} l c c c }
        \toprule
        Method & MMD ($\times$ $10^3$) $\downarrow$ & 1-NAA $\downarrow$ & COV $\uparrow$ \\
        \midrule
        Coordinates  & 19.26 & 71.71 & 47.56 \\
        Index Offset & \textbf{15.26} & \textbf{50.73} & \textbf{58.29} \\
        \bottomrule
    \end{tabular*}
\end{table}

\begin{table}[t]
  \centering
  \small
  
  \caption{
    Memory (GB) and runtime (s) comparison w/o NSCA.
    Each configuration is evaluated and averaged over 128 iterations. 
    Context and window sizes are measured in number of faces.
  }
  \label{tab:nsca_performance}
  
  \begin{tabular}{c c c c c c}
    \hline
    \multirow{2}{*}[-0.8ex]{Total Faces} & 
    \multirow{2}{*}[-0.8ex]{Window} & 
    \multicolumn{2}{c}{\multirow{2}{*}[0.8ex]{NSCA}} & 
    \multicolumn{2}{c}{\multirow{2}{*}[0.8ex]{w/o NSCA}} \\
    \cmidrule(lr){3-4} \cmidrule(lr){5-6}
    & & \makecell{Time} & \makecell{Mem} & \makecell{Time} & \makecell{Mem} \\
    \hline
    5K & 1K & 0.253 & 24.65 & 0.180 & 27.68 \\
    10K & 1K & 0.252 & 24.72 & 0.199 & 28.50 \\
    20K & 2K & 0.398 & 30.16 & 0.419 & 38.15 \\
    50K & 5K & 1.229 & 48.46 & \multicolumn{2}{c}{OOM} \\
    100K & 5K & 1.275 & 49.19 & \multicolumn{2}{c}{OOM} \\
    \hline
  \end{tabular}
  \vspace{-1em}
\end{table}




    

\vspace{-1em}
\subsection{Ablation Study}
\noindent \textbf{Modules.} We conduct ablation studies to quantify the effect of each proposed component. Specifically, we remove the frontier mask, context injection, the NSCA module, non-manifold encoding, and expansive reasoning in turn, and evaluate on 5k-face meshes with a validation subset of 128 meshes. Performance of CD and HD is shown in \Cref{tab:ablation_subtractive}. The use of full context without NSCA achieves the best performance, while NSCA significant saves memory and time with minor performance drop.

\noindent \textbf{Root Prediction.} We further ablate the next-root prediction schemes: directly predict vertex coordinates versus predicting a relative offset step, as shown in \Cref{tab:ablation_root_pred}. 

\noindent \textbf{Resources w/o NSCA.} \Cref{tab:nsca_performance} compares the computational resource consumption with and without NSCA across various configurations. We set the NSCA kernel size to 32, sliding stride to 16, compression block size to 64, and select top 16 blocks.

%

%% file: sec/5_conclusion.tex
\vspace{-0.5em}
\section{Conclusion}
\vspace{-0.5em}
We introduce \textbf{MeshRipple}, a structured autoregressive framework for artist-mesh generation with a frontier-aware BFS tokenization, an expansive prediction scheme, and a lightweight sparse contextual attention. It's compatible with truncated training and produces complete, high-quality meshes, outperforming strong recent baselines.

\noindent \textbf{Limitations and future work.} Substantial label noise or extreme configurations can degrade our model's performance. We plan to increase coordinate quantization resolution and scale to meshes with up to 100K faces, extend the framework to quad meshes, and merge our model with RL-based preference tuning.

%

%% file: supplementary_material/1_details_of_tokenization_algorithm.tex
\section{Ripple Tokenization Algorithm}
\RestyleAlgo{ruled}
\begin{algorithm}
\setstretch{1.1}
\caption{Ripple Tokenization}\label{alg:bfs_tok}
\SetStartEndCondition{ }{}{}%
\SetKwProg{Fn}{def}{\string:}{}
\SetKw{KwTo}{in}\SetKwFor{For}{for}{\string:}{}%
\SetKwIF{If}{ElseIf}{Else}{if}{:}{elif}{else:}{}%
\SetKwFor{While}{while}{:}{fintq}%
\AlgoDontDisplayBlockMarkers\SetAlgoNoEnd\SetAlgoNoLine%
\SetKwComment{Comment}{/* }{ */}
\SetKwInOut{Input}{Input}
\SetKwInOut{Output}{Output}
\SetKwInOut{Preprocess}{Preprocess}

\Input{Mesh $\mathcal{M}$}
\Preprocess{$\mathcal{M}$ $\rightarrow$ \,\, \text{Half Edges} \,\, $\mathbf H = \{\mathbf{h}_i\}_N$ }
\Output{Sequence $\mathcal{S}$, Root Index $\mathbf{R}$}
\KwData{FIFO Frontier Queue $\mathcal{B}$}

\BlankLine
\SetKwFunction{RecordRoot}{RecordRoot}%
\SetKwFunction{AddFace}{AddFace}%
$\mathcal{S}$.append(\texttt{BOS})\;
\For{$\mathrm{HalfEdge}$ $\mathbf{h}$ in H}{
    \If{$\mathbf{h}.\mathbf{face}.vis$}{\textbf{continue}\;}
    $\mathbf{h}.\mathbf{face}.vis = \text{true}$\;
    $\mathcal{B}$.enqueue($\mathbf{h}$)\;
    $\mathcal{S}$.append(\texttt{N})\;
    \AddFace($\mathcal{S}$, $\mathbf{h}$)\;
    \While{$\mathcal{B}$ $\mathrm{is\ not\ empty}$}{
        $\mathbf{h} = \mathcal{B}$.dequeue()\;
        \For{$\mathrm{HalfEdge}$ $\mathbf{h}_o$ in $[ \mathbf{h}.\mathbf{p}.\mathbf{o}; \mathbf{h}.\mathbf{n}.\mathbf{o}; \mathbf{h}.\mathbf{o}]$}{
            \If{not $\mathbf{h}_o.\mathbf{face}.vis$}{
                $\mathbf{h}_o.\mathbf{face}.vis = \text{true}$\;
                \AddFace($\mathcal{S}$, $\mathbf{h}_o$)\; 
                \RecordRoot($\mathbf{R}$, $\mathbf{h}$, $\mathbf{h}_o$)\;
                $\mathcal{B}$.enqueue($\mathbf{h}_o$);
            }
        }
    }
}
Project $\mathcal{S}$ to vertex coordinates

Output $\mathcal{S}$, $\mathbf{R}$
\end{algorithm}
In this section, we elaborate on the proposed Ripple Tokenization. We begin by preprocessing the input mesh: vertices are quantized into discrete bins (set to 256 in our experiments), duplicated vertices are merged, and degenerate faces are removed. This yields a sanitized mesh $\mathcal{M}=\{ \mathcal{F}, \mathcal{V}\} $, where each face $\mathbf{F}_i \in \mathcal{F}$ is defined as $\mathbf{F}_i = [\mathbf{a}^i_0, \mathbf{a}^i_1, \mathbf{a}^i_2]$, containing the indices of the vertices forming the face.
To establish a deterministic traversal order, we sort all faces lexicographically based on their vertex coordinates (specifically using $z$-$y$-$x$ ordering). Subsequently, we construct a half-edge data structure for the mesh, respecting the orientation defined by the face normals.
The tokenization process employs a breadth-first traversal initialized at the first half-edge of the lexicographically first face, $\mathbf{F}_0$. During each iteration, a half-edge $\mathbf{h}$ is dequeued, and we sequentially access the twin half-edges of its predecessor ($\mathbf{h}.p$), its successor ($\mathbf{h}.n$), and the half-edge itself ($\mathbf{h}$). This sequence strictly defines the traversal priority for the neighbors of any given face $\mathbf{F}_i$.
To prevent cycles and redundancy, we maintain a binary visitation state, $\mathtt{vis}$, for each face. When a neighboring face is accessed via a twin half-edge, we check its state; if it has not yet been visited, we mark $\mathtt{vis}$ as true and add the corresponding half-edge to the queue. This propagation continues until all reachable half-edges in the connected component have been processed.

\begin{figure}[h] 
\centering 
\includegraphics[width=0.7\columnwidth]{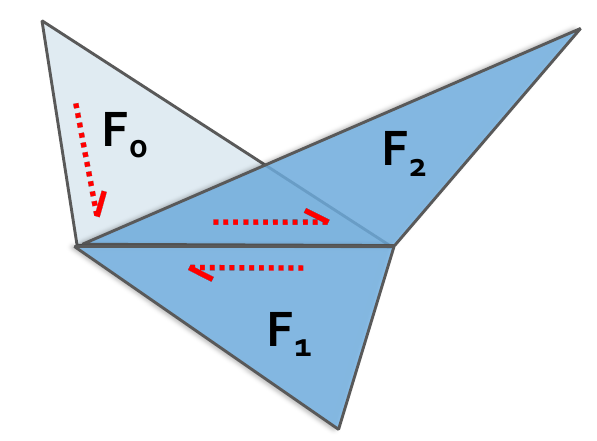} 
\caption{\textbf{Illustration of Non-Manifold Traversal Strategy.} Direct traversal between $F_0$ and $F_2$ is inhibited because their incident half-edges are co-directional. The algorithm resolves this by extending traversal through the intermediate face $F_1$, maintaining normal vector consistency.}
\label{fig:nonmanifold_tokenizer}
\vspace{-1.5em}
\end{figure}

\noindent \textbf{Non-manifoldness.}
To ensure consistent surface orientation, our algorithm restricts traversal across non-manifold edges to incident half-edges with opposite orientations. Consequently, traversal is explicitly inhibited when an adjacent half-edge is co-directional with the current half-edge. As illustrated in \Cref{fig:nonmanifold_tokenizer}, although faces $\mathbf{F}_0$ and $\mathbf{F}_2$ share identical normal vectors, the co-directionality of their shared boundary prevents direct propagation between them.
However, structural connectivity allows for indirect traversal. In the example shown, the algorithm successfully propagates from $\mathbf{F}_0$ to $\mathbf{F}_1$, provided their interface satisfies the orientation constraint (opposite half-edges). From $\mathbf{F}_1$, the traversal can subsequently extend to $\mathbf{F}_2$. Crucially, in scenarios where a face is effectively isolated by co-directional half-edges on all sides---indicating an irresolvable conflict in face normals---we interpret this as a topological discontinuity and segregate the regions into separate connected components.

%

\begin{figure}[h] 
\centering 
\includegraphics[width=\columnwidth]{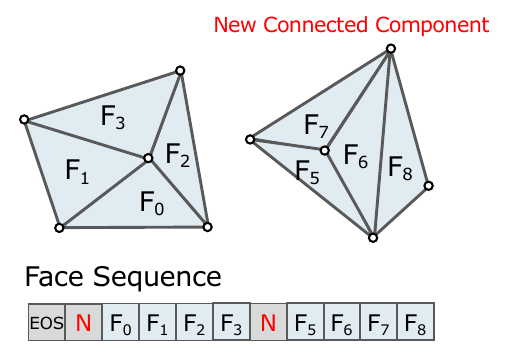} 
\caption{\textbf{Initialization of Disconnected Components.} When initiating traversal for a topologically disjoint component, the starting face lacks a preceding frontier face. A special identifier $N$ is used to explicitly mark this boundary.}
\label{fig:new_connected_component_tokenizer}
\vspace{-1.5em}
\end{figure}
\noindent \textbf{Disconnected Components.}
For each independent connected component, the starting face lacks a preceding frontier face. Consequently, we assign a special token \colorbox{GreenYellow}{\textbf{N}} to explicitly mark the initiation of a new component, as illustrated in \Cref{fig:new_connected_component_tokenizer}.

%% file: supplementary_material/2_more_implementation_details.tex
\section{More Implementation Details}

\subsection{Filtering Training Data}
\begin{table}[h]
    \centering
    \caption{\textbf{Model Architecture Hyperparameters.}}
    \label{tab:model_params}
    
    \begin{tabular}{lcc}
        \toprule
         & \textbf{Small scale} & \textbf{Large scale} \\
        \midrule
        Parameter count & 500 M & 800M \\
        Batch Size & 8 & 4 \\
        Hourglass & 4,4,9 & 4,4,9 \\
        Heads & 10 & 16 \\
        $d_{model}$ & 768 & 1024 \\
        $d_{FFN}$ & 3072 & 4096 \\
        Learning rate & $1e-4$ & $1e-4$ \\
        LR scheduler & Cosine & Cosine \\
        Weight decay & 0.1 & 0.1 \\
        Gradient Clip & 1.0 & 1.0 \\
        \bottomrule
    \end{tabular}
    \label{model_params}
\end{table}
\subsection{More Training Details}
The significant variance in data quality across collected datasets poses a challenge for robust model training. To address this, we implement a strict filtering pipeline based on geometric topology and quantization artifacts.

\noindent \textbf{Geometric and Topological Filtering.}
We discard meshes with face counts outside the range $[500, 20\text{k}]$, vertex-to-face ratios $>0.8$, or where narrow faces (minimum angle $<5^\circ$) exceed $20\%$ of the total. To ensure topological simplicity, we exclude meshes exceeding a maximum BFS displacement of $100$, a boundary length of $500$, or $20$ connected components. The component count is determined after pruning small clusters ($<10$ faces) located within a normalized distance of $0.05$ from valid components.

\noindent \textbf{Vertex Merge Filtering.}
To mitigate artifacts from discrete vertex merging, we filter meshes where the vertex count drops by $>50\%$, or where newly introduced self-intersecting or overlapping faces exceed $10\%$. Self-intersections are detected via \texttt{trimesh}, while overlapping faces are identified by grouping coplanar faces and verifying candidates via the Separating Axis Theorem (SAT).

We implemented two variants of MeshRipple: a small-scale model and a large-scale model. 
The detailed architectural specifications are presented in Table \ref{tab:model_params}. 
It is worth noting that the reported parameter counts include the parameters of Michelangelo encoder. 
Furthermore, to optimize GPU memory usage, we utilized bfloat16 (bf16) precision and FlashAttention.

Furthermore, we employ point cloud noise injection, rotation, and scaling for data augmentation.
Specifically, during point cloud sampling, Gaussian noise is added to the vertex coordinates with a probability of $0.5$. 
For rotation, the mesh is set to perform rotational augmentation around the $z$-axis, with each unit being $30^\circ$, covering a total of $360^\circ$. 
Finally, for scaling, we uniformly sample a scaling factor from the range $[0.75, 1.25]$ for each axis.

%% file: supplementary_material/3_more_ablation_study.tex
\section{More Ablation Study}
\begin{table}[ht]
  \centering
  \caption{\textbf{Ablation study on the truncation window size.}
  Metrics are evaluated on 160 dense meshes.}
  \label{tab:window_ablation} 
  \begin{tabular}{lcccc}
    \toprule
    Window Size & CD ($\downarrow$) & HD ($\downarrow$) & NC ($\uparrow$) \\
    \midrule
    1k & 0.051948 & 0.109130 & 0.796585 \\
    2k & \textbf{0.050651} & \textbf{0.104539} & \textbf{0.798088} \\
    \bottomrule
  \end{tabular}
\end{table}

\begin{table}[h]
  \centering
  \small
  
  \caption{
    \textbf{GPU memory (GB) and runtime (s) comparison w/o NSCA on a single A800 GPU.}
    Each configuration is evaluated and averaged over 128 iterations.
    Context and window sizes are measured in number of faces.
  }
  \label{tab:nsca_performance}
  
  \begin{tabular}{c c c c c c}
    \hline
    \multirow{2}{*}[-0.8ex]{Total Faces} & 
    \multirow{2}{*}[-0.8ex]{Window} & 
    \multicolumn{2}{c}{\multirow{2}{*}[0.8ex]{NSCA}} & 
    \multicolumn{2}{c}{\multirow{2}{*}[0.8ex]{w/o NSCA}} \\
    \cmidrule(lr){3-4} \cmidrule(lr){5-6}
    & & \makecell{Time} & \makecell{Mem} & \makecell{Time} & \makecell{Mem} \\
    \hline
    5K & 1K & 0.253 & 24.65 & 0.180 & 27.68 \\
    5K & 2K & 0.391 & 29.94 & 0.335 & 33.68 \\
    5K & 5K & 1.199 & 47.85 & 1.156 & 53.59 \\
    10K & 1K & 0.252 & 24.72 & 0.199 & 28.50 \\
    10K & 2K & 0.393 & 30.01 & 0.362 & 35.17 \\
    10K & 5K & 1.194 & 47.91 & 1.207 & 57.13 \\
    20K & 1K & 0.254 & 24.86 & 0.244 & 30.16 \\
    20K & 2K & 0.398 & 30.16 & 0.419 & 38.15 \\
    20K & 5K & 1.200 & 48.05 & \multicolumn{2}{c}{OOM} \\
    50K & 1K & 0.265 & 25.34 & 0.426 & 35.09 \\
    50K & 2K & 0.420 & 30.61 & 0.636 & 47.19 \\
    50K & 5K & 1.229 & 48.46 & \multicolumn{2}{c}{OOM} \\
    100K & 1K & 0.300 & 26.13 & 0.726 & 43.35 \\
    100K & 2K & 0.458 & 31.42 & 0.984 & 62.26 \\
    100K & 5K & 1.275 & 49.19 & \multicolumn{2}{c}{OOM} \\
    \hline
  \end{tabular}
\end{table}

\noindent \textbf{Truncation Window Size.}
We conduct an ablation study on the truncation window size, comparing 1k and 2k, on a dataset of 100k samples. The evaluation is performed on 160 dense meshes, with results presented in Table \ref{tab:window_ablation}.

\noindent \textbf{Resources w/o NSCA.}
\Cref{tab:nsca_performance} compares the computational resource consumption with and without NSCA across various configurations. We have noticed the advantages of NSCA in terms of time and GPU memory resource consumption in long sequences, which can help us extend MeshRipple to longer window and longer mesh face amount.

%% file: supplementary_material/5_more_results.tex
\section{More Results}
\begin{figure*}[h] 
\centering 
\includegraphics[width=0.9\textwidth]{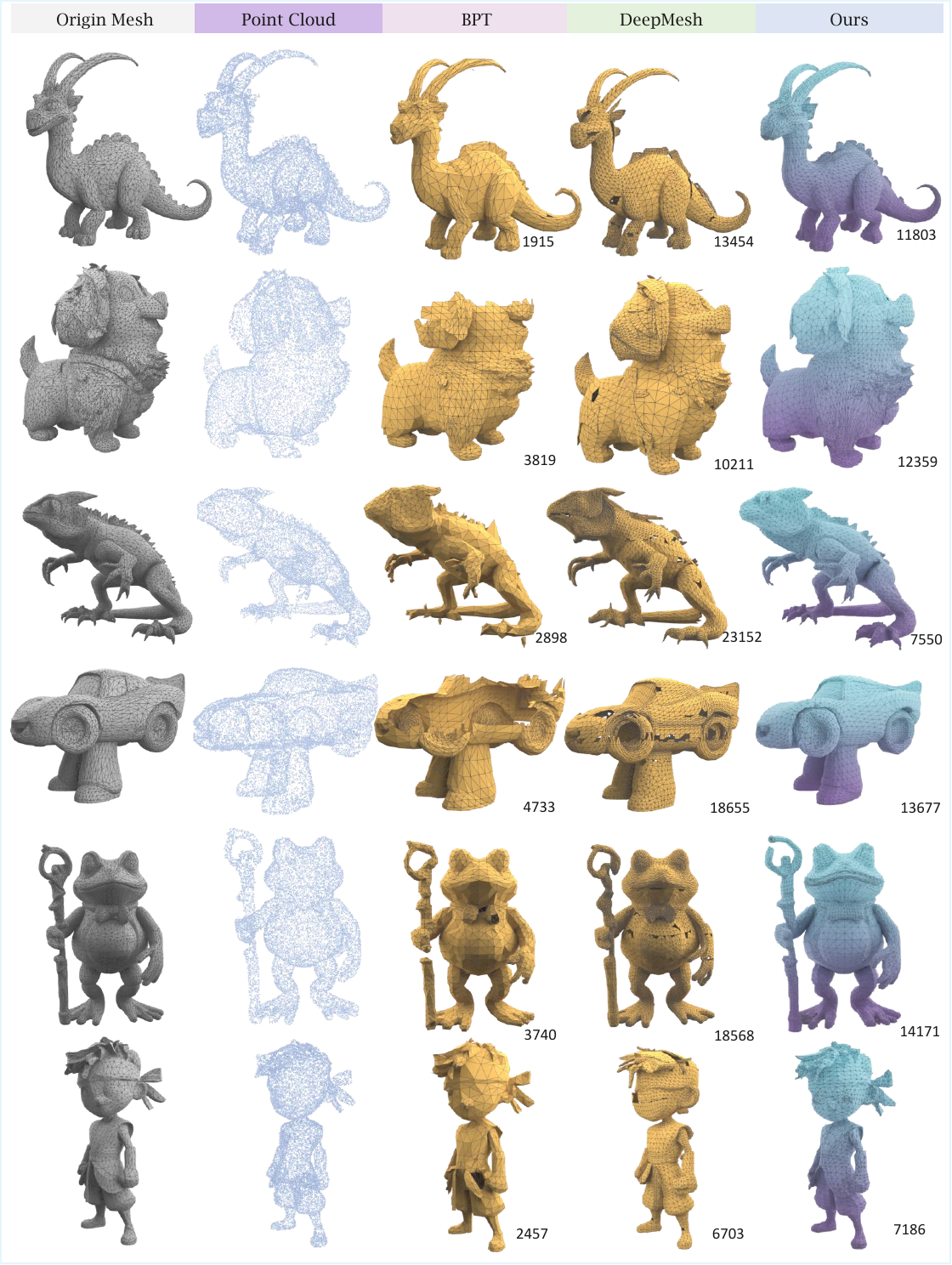} 
\caption{\textbf{More comparisons between MeshRipple and the SOTAs.}}
\label{fig:more_comp}
\end{figure*}

\begin{figure*}[h] 
\centering 
\includegraphics[width=0.85\textwidth]{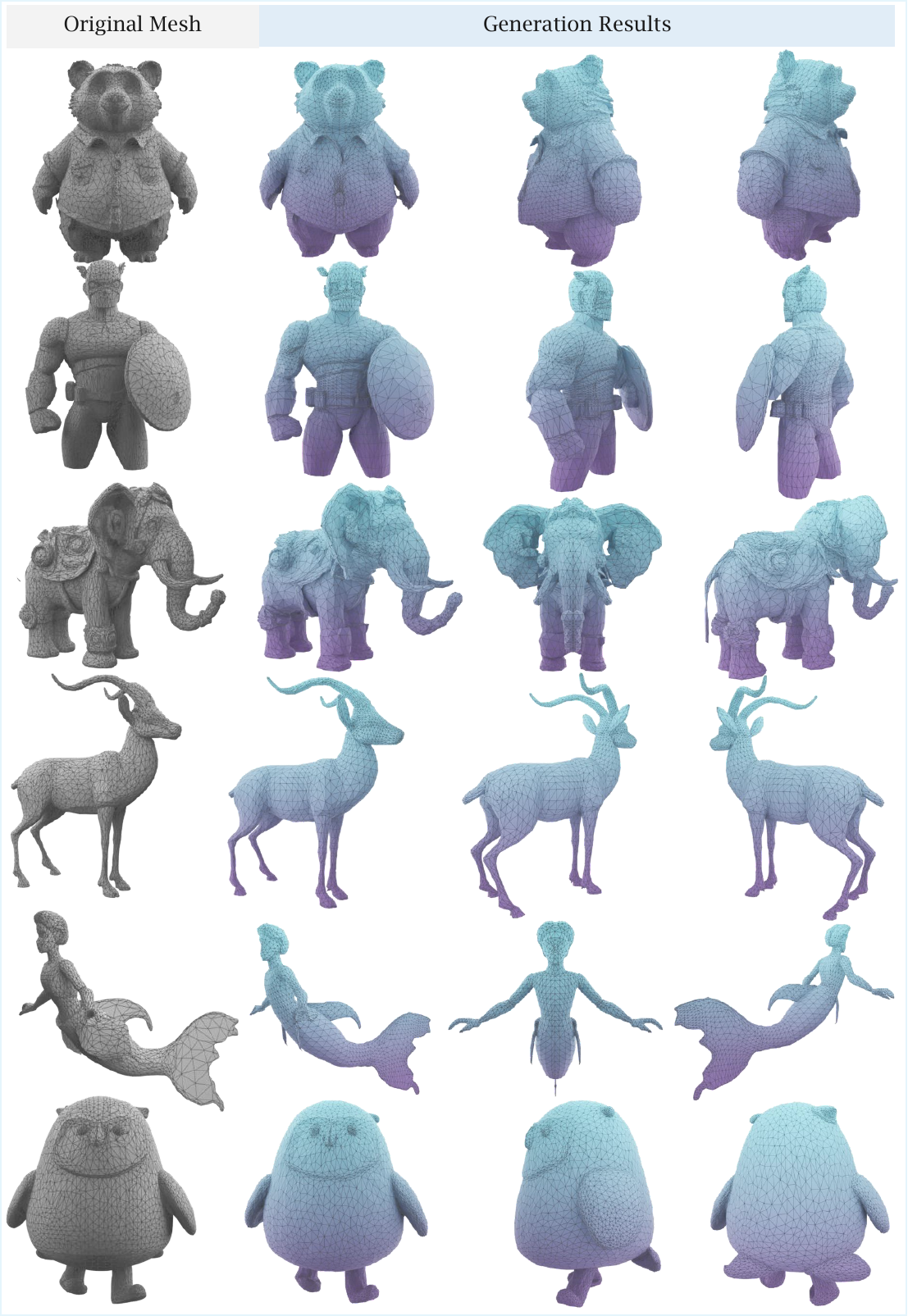} 
\caption{\textbf{More results generated by MeshRipple.}}
\label{fig:more_res_p1}
\end{figure*}

\begin{figure*}[h] 
\centering 
\includegraphics[width=0.85\textwidth]{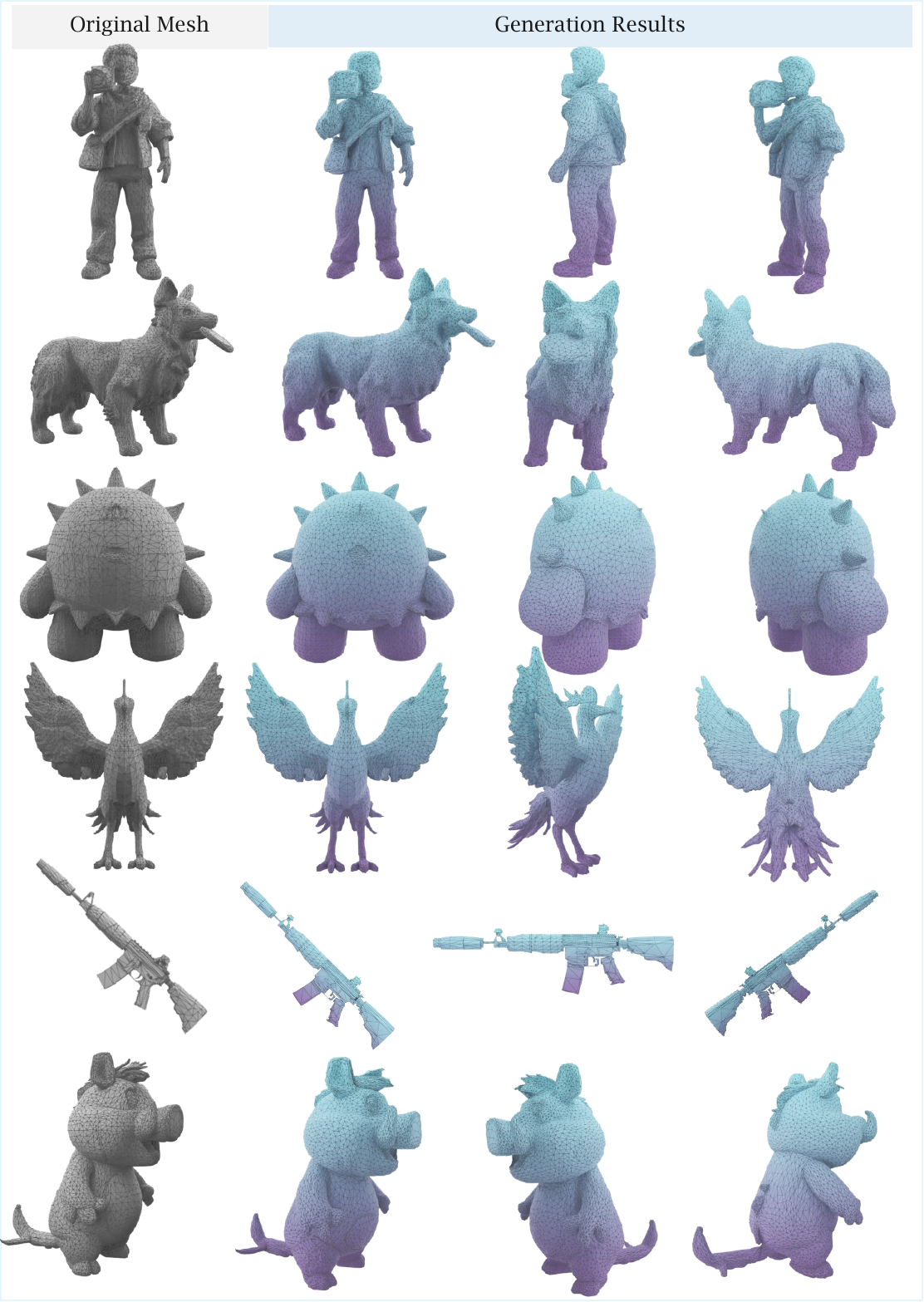} 
\caption{\textbf{More results generated by MeshRipple.}}
\label{fig:more_res_p2}
\end{figure*}

\begin{figure*}[h] 
\centering 
\includegraphics[width=1\textwidth]{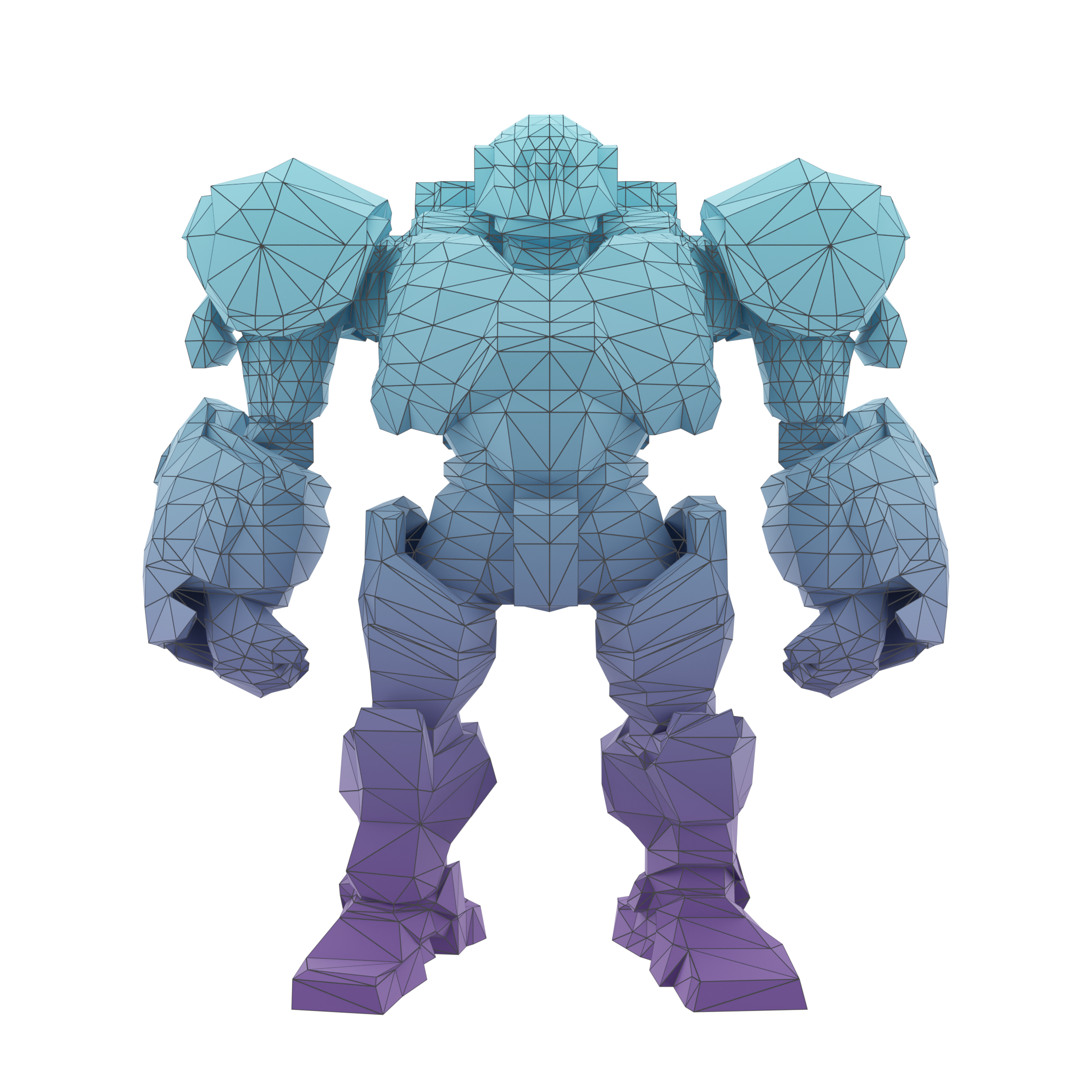} 
\caption{\textbf{High-fidelity visualizations of generated meshes.}}
\label{fig:high-resolution_res1}
\end{figure*}

\begin{figure*}[h] 
\centering 
\includegraphics[width=1\textwidth]{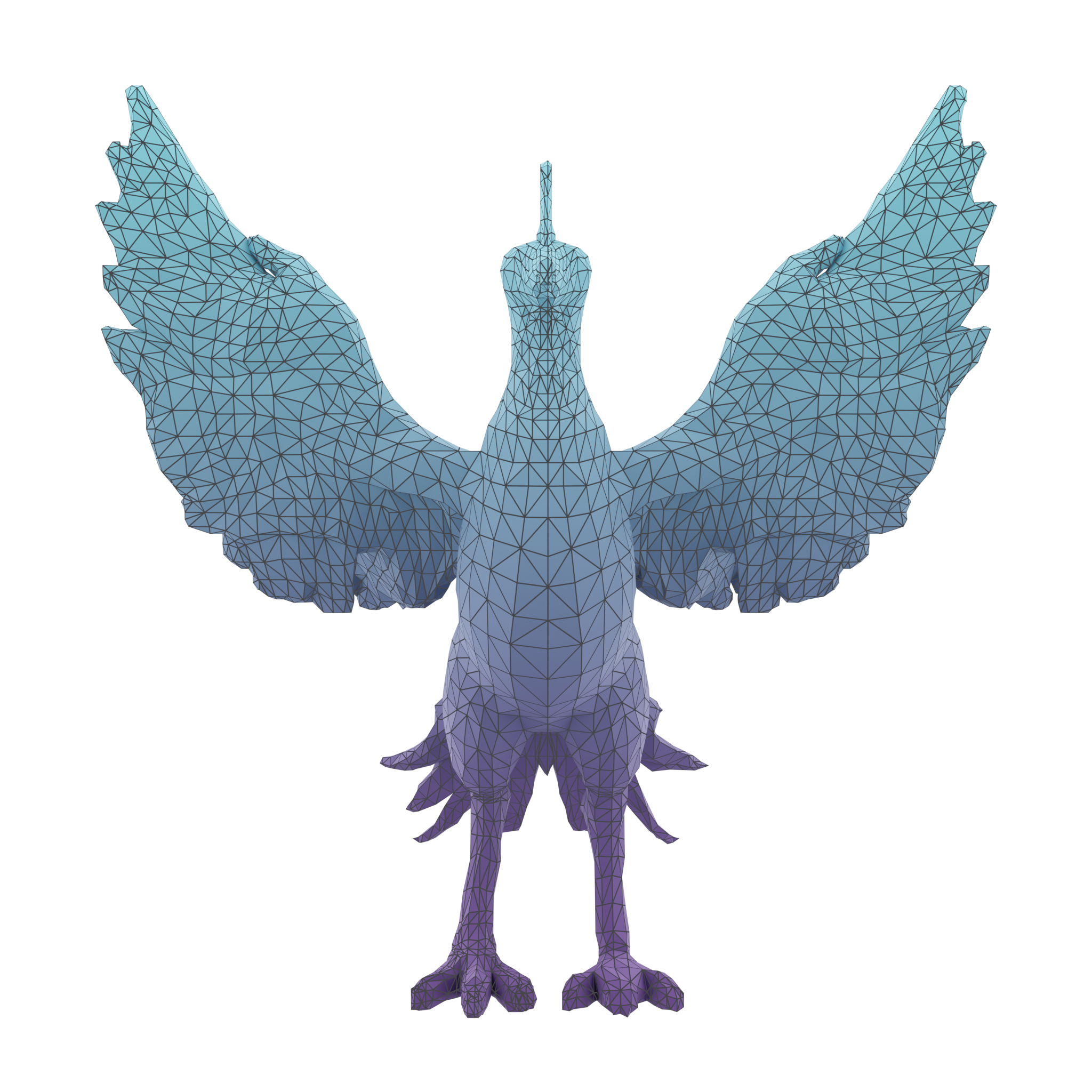} 
\caption{\textbf{High-fidelity visualizations of generated meshes.}}
\label{fig:high-resolution_res2}
\end{figure*}

\begin{figure*}[h] 
\centering 
\includegraphics[width=1\textwidth]{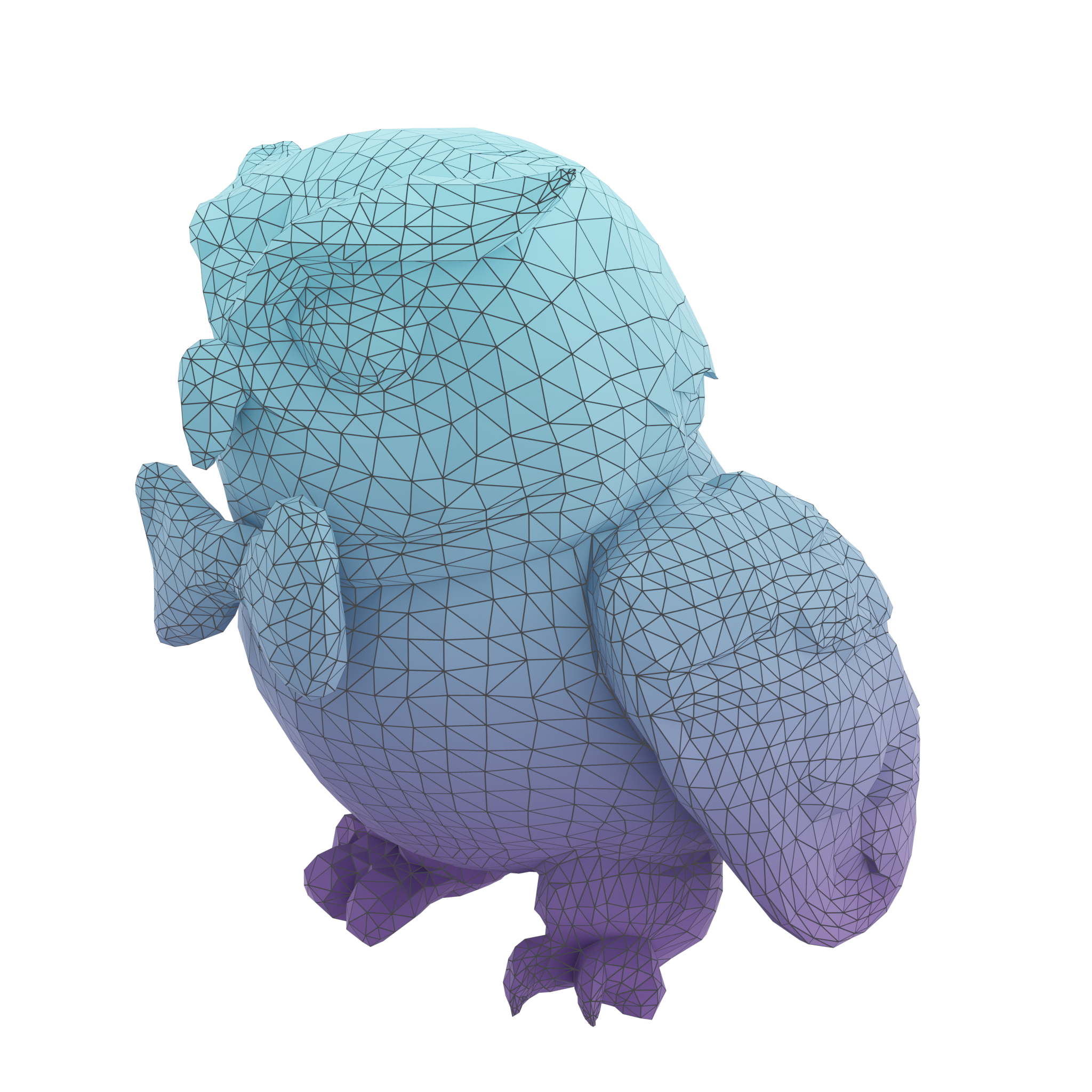} 
\caption{\textbf{High-fidelity visualizations of generated meshes.}}
\label{fig:high-resolution_res3}
\end{figure*}

\begin{figure*}[h] 
\centering 
\includegraphics[width=1\textwidth]{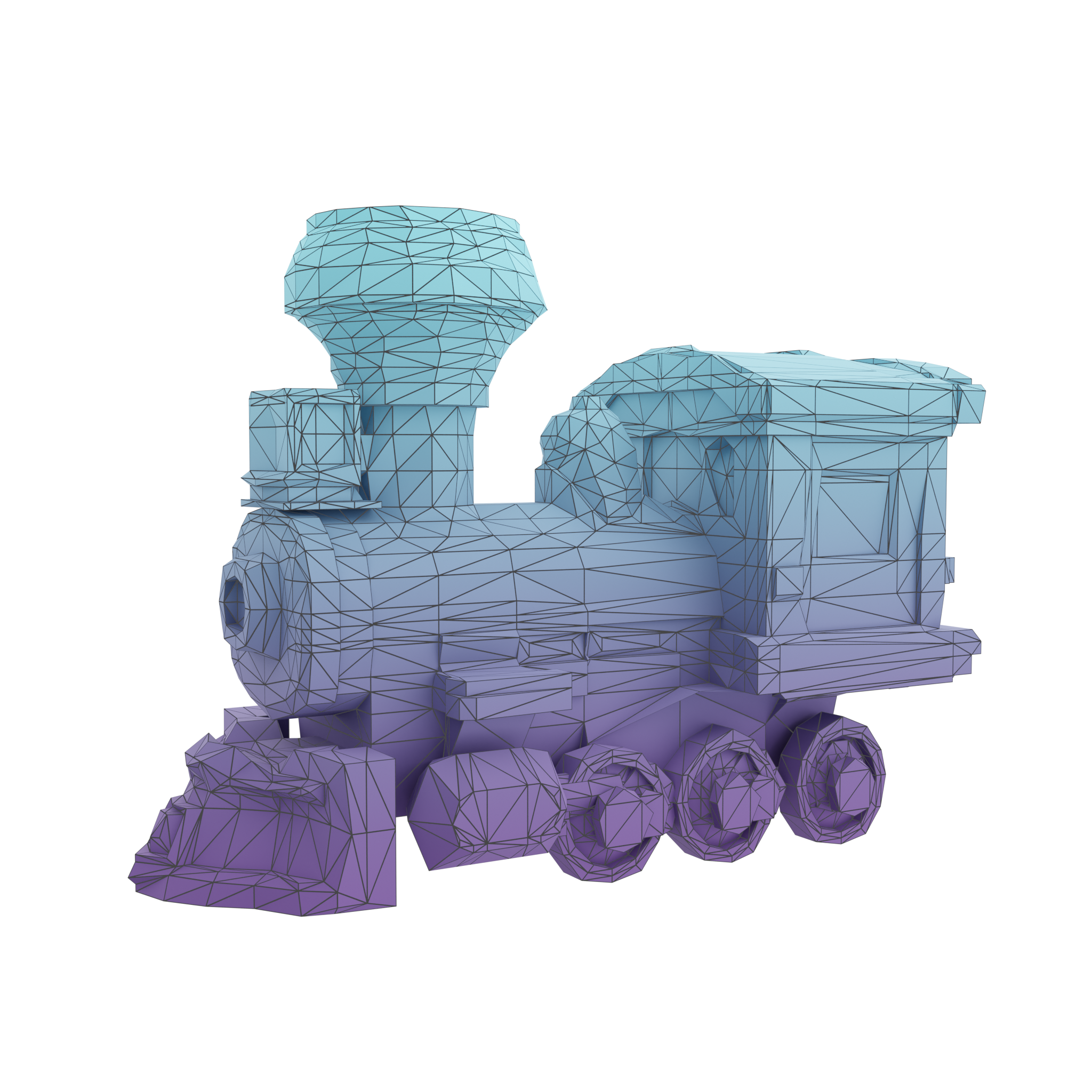} 
\caption{\textbf{High-fidelity visualizations of generated meshes.}}
\label{fig:high-resolution_res4}
\end{figure*}

We present additional qualitative comparisons with BPT and DeepMesh in \Cref{fig:more_comp}, where the face count for each mesh is explicitly annotated. Furthermore, extended results for point cloud-conditioned generation are illustrated in \Cref{fig:more_res_p1} and \Cref{fig:more_res_p2}. Finally, \Cref{fig:high-resolution_res1}, \Cref{fig:high-resolution_res2}, \Cref{fig:high-resolution_res3} and \Cref{fig:high-resolution_res4} displays high-resolution visualizations to facilitate a detailed inspection of the geometric fine structures.

%% file: main.bib
@String(CVPR= {IEEE Conf. Comput. Vis. Pattern Recog.})

@String(ICCV= {Int. Conf. Comput. Vis.})

@String(ECCV= {Eur. Conf. Comput. Vis.})

@String(ICLR = {Int. Conf. Learn. Represent.})

@String(CVPR  = {CVPR})

@String(ICCV  = {ICCV})

@String(ECCV  = {ECCV})

@String(ICLR  = {ICLR})

@inproceedings{siddiqui2024meshgpt,
  author       = {Yawar Siddiqui and
                  Antonio Alliegro and
                  Alexey Artemov and
                  Tatiana Tommasi and
                  Daniele Sirigatti and
                  Vladislav Rosov and
                  Angela Dai and
                  Matthias Nie{\ss}ner},
  title        = {MeshGPT: Generating Triangle Meshes with Decoder-Only Transformers},
  booktitle    = {{IEEE/CVF} Conference on Computer Vision and Pattern Recognition,
                  {CVPR} 2024, Seattle, WA, USA, June 16-22, 2024},
  pages        = {19615--19625},
  publisher    = {{IEEE}},
  year         = {2024},
  url          = {https://doi.org/10.1109/CVPR52733.2024.01855},
  doi          = {10.1109/CVPR52733.2024.01855},
  bibsource    = {dblp computer science bibliography, https://dblp.org}
}

@inproceedings{nash2020polygen,
  author       = {Charlie Nash and
                  Yaroslav Ganin and
                  S. M. Ali Eslami and
                  Peter W. Battaglia},
  title        = {PolyGen: An Autoregressive Generative Model of 3D Meshes},
  booktitle    = {Proceedings of the 37th International Conference on Machine Learning,
                  {ICML} 2020, 13-18 July 2020, Virtual Event},
  series       = {Proceedings of Machine Learning Research},
  volume       = {119},
  pages        = {7220--7229},
  publisher    = {{PMLR}},
  year         = {2020},
  url          = {http://proceedings.mlr.press/v119/nash20a.html},
  bibsource    = {dblp computer science bibliography, https://dblp.org}
}

@inproceedings{lionar2025treemeshgpt,
  title={Treemeshgpt: Artistic mesh generation with autoregressive tree sequencing},
  author={Lionar, Stefan and Liang, Jiabin and Lee, Gim Hee},
  booktitle={Proceedings of the Computer Vision and Pattern Recognition Conference},
  pages={26608--26617},
  year={2025}
}

@article{song2025mesh,
  title={Mesh Silksong: Auto-Regressive Mesh Generation as Weaving Silk},
  author={Song, Gaochao and Zhao, Zibo and Weng, Haohan and Zeng, Jingbo and Jia, Rongfei and Gao, Shenghua},
  journal={arXiv preprint arXiv:2507.02477},
  year={2025}
}

@article{chen2024meshxl,
  title={Meshxl: Neural coordinate field for generative 3d foundation models},
  author={Chen, Sijin and Chen, Xin and Pang, Anqi and Zeng, Xianfang and Cheng, Wei and Fu, Yijun and Yin, Fukun and Wang, Billzb and Yu, Jingyi and Yu, Gang and others},
  journal={Advances in Neural Information Processing Systems},
  volume={37},
  pages={97141--97166},
  year={2024}
}

@article{hao2024meshtron,
  title={Meshtron: High-fidelity, artist-like 3d mesh generation at scale},
  author={Hao, Zekun and Romero, David W and Lin, Tsung-Yi and Liu, Ming-Yu},
  journal={arXiv preprint arXiv:2412.09548},
  year={2024}
}

@article{kim2025fastmesh,
  title={FastMesh: Efficient Artistic Mesh Generation via Component Decoupling},
  author={Kim, Jeonghwan and Lan, Yushi and Fortes, Armando and Chen, Yongwei and Pan, Xingang},
  journal={arXiv preprint arXiv:2508.19188},
  year={2025}
}

@inproceedings{chen2024meshanything,
  author       = {Yiwen Chen and
                  Tong He and
                  Di Huang and
                  Weicai Ye and
                  Sijin Chen and
                  Jiaxiang Tang and
                  Zhongang Cai and
                  Lei Yang and
                  Gang Yu and
                  Guosheng Lin and
                  Chi Zhang},
  title        = {MeshAnything: Artist-Created Mesh Generation with Autoregressive Transformers},
  booktitle    = {The Thirteenth International Conference on Learning Representations,
                  {ICLR} 2025, Singapore, April 24-28, 2025},
  publisher    = {OpenReview.net},
  year         = {2025},
  url          = {https://openreview.net/forum?id=KGZAs8VcOM},
  bibsource    = {dblp computer science bibliography, https://dblp.org}
}

@inproceedings{chen2025meshanything,
  title={Meshanything v2: Artist-created mesh generation with adjacent mesh tokenization},
  author={Chen, Yiwen and Wang, Yikai and Luo, Yihao and Wang, Zhengyi and Chen, Zilong and Zhu, Jun and Zhang, Chi and Lin, Guosheng},
  booktitle={Proceedings of the IEEE/CVF International Conference on Computer Vision},
  pages={13922--13931},
  year={2025}
}

@inproceedings{tang2024edgerunner,
  author       = {Jiaxiang Tang and
                  Zhaoshuo Li and
                  Zekun Hao and
                  Xian Liu and
                  Gang Zeng and
                  Ming{-}Yu Liu and
                  Qinsheng Zhang},
  title        = {EdgeRunner: Auto-regressive Auto-encoder for Artistic Mesh Generation},
  booktitle    = {The Thirteenth International Conference on Learning Representations,
                  {ICLR} 2025, Singapore, April 24-28, 2025},
  publisher    = {OpenReview.net},
  year         = {2025},
  url          = {https://openreview.net/forum?id=81cta3WQVI},
  bibsource    = {dblp computer science bibliography, https://dblp.org}
}

@article{rossignac2002edgebreaker,
  title={Edgebreaker: Connectivity compression for triangle meshes},
  author={Rossignac, Jarek},
  journal={IEEE transactions on visualization and computer graphics},
  volume={5},
  number={1},
  pages={47--61},
  year={2002},
  publisher={IEEE}
}

@inproceedings{zhao2025deepmesh,
  title={Deepmesh: Auto-regressive artist-mesh creation with reinforcement learning},
  author={Zhao, Ruowen and Ye, Junliang and Wang, Zhengyi and Liu, Guangce and Chen, Yiwen and Wang, Yikai and Zhu, Jun},
  booktitle={Proceedings of the IEEE/CVF International Conference on Computer Vision},
  pages={10612--10623},
  year={2025}
}

@article{chang2015shapenetinformationrich3dmodel,
  title={Shapenet: An information-rich 3d model repository},
  author={Chang, Angel X and Funkhouser, Thomas and Guibas, Leonidas and Hanrahan, Pat and Huang, Qixing and Li, Zimo and Savarese, Silvio and Savva, Manolis and Song, Shuran and Su, Hao and others},
  journal={arXiv preprint arXiv:1512.03012},
  year={2015}
}

@inproceedings{zuo2024sparse3d,
     title={High-Fidelity 3D Textured Shapes Generation by Sparse Encoding and Adversarial Decoding},
     author={Zuo, Qi and Gu, Xiaodong and Dong, Yuan and Zhao, Zhengyi and Yuan, Weihao and Qiu, Lingteng and Bo, Liefeng and Dong, Zilong},
     booktitle={European Conference on Computer Vision},
     year={2024}
     }

@inproceedings{Richdreamer,
  author       = {Lingteng Qiu and
                  Guanying Chen and
                  Xiaodong Gu and
                  Qi Zuo and
                  Mutian Xu and
                  Yushuang Wu and
                  Weihao Yuan and
                  Zilong Dong and
                  Liefeng Bo and
                  Xiaoguang Han},
  title        = {RichDreamer: {A} Generalizable Normal-Depth Diffusion Model for Detail
                  Richness in Text-to-3D},
  booktitle    = {{IEEE/CVF} Conference on Computer Vision and Pattern Recognition,
                  {CVPR} 2024, Seattle, WA, USA, June 16-22, 2024},
  pages        = {9914--9925},
  publisher    = {{IEEE}},
  year         = {2024},
  url          = {https://doi.org/10.1109/CVPR52733.2024.00946},
  doi          = {10.1109/CVPR52733.2024.00946},
  bibsource    = {dblp computer science bibliography, https://dblp.org}
}

@inproceedings{stojanov2021using,
  title={Using shape to categorize: Low-shot learning with an explicit shape bias},
  author={Stojanov, Stefan and Thai, Anh and Rehg, James M},
  booktitle={Proceedings of the IEEE/CVF conference on computer vision and pattern recognition},
  pages={1798--1808},
  year={2021}
}

@inproceedings{Lrm,
  author       = {Yicong Hong and
                  Kai Zhang and
                  Jiuxiang Gu and
                  Sai Bi and
                  Yang Zhou and
                  Difan Liu and
                  Feng Liu and
                  Kalyan Sunkavalli and
                  Trung Bui and
                  Hao Tan},
  title        = {{LRM:} Large Reconstruction Model for Single Image to 3D},
  booktitle    = {The Twelfth International Conference on Learning Representations,
                  {ICLR} 2024, Vienna, Austria, May 7-11, 2024},
  publisher    = {OpenReview.net},
  year         = {2024},
  url          = {https://openreview.net/forum?id=sllU8vvsFF},
  bibsource    = {dblp computer science bibliography, https://dblp.org}
}

@inproceedings{Instant3d,
  author       = {Jiahao Li and
                  Hao Tan and
                  Kai Zhang and
                  Zexiang Xu and
                  Fujun Luan and
                  Yinghao Xu and
                  Yicong Hong and
                  Kalyan Sunkavalli and
                  Greg Shakhnarovich and
                  Sai Bi},
  title        = {Instant3D: Fast Text-to-3D with Sparse-view Generation and Large Reconstruction
                  Model},
  booktitle    = {The Twelfth International Conference on Learning Representations,
                  {ICLR} 2024, Vienna, Austria, May 7-11, 2024},
  publisher    = {OpenReview.net},
  year         = {2024},
  url          = {https://openreview.net/forum?id=2lDQLiH1W4},
  bibsource    = {dblp computer science bibliography, https://dblp.org}
}

@inproceedings{poole2022dreamfusion,
  author       = {Ben Poole and
                  Ajay Jain and
                  Jonathan T. Barron and
                  Ben Mildenhall},
  title        = {DreamFusion: Text-to-3D using 2D Diffusion},
  booktitle    = {The Eleventh International Conference on Learning Representations,
                  {ICLR} 2023, Kigali, Rwanda, May 1-5, 2023},
  publisher    = {OpenReview.net},
  year         = {2023},
  url          = {https://openreview.net/forum?id=FjNys5c7VyY},
  bibsource    = {dblp computer science bibliography, https://dblp.org}
}

@incollection{marchingcube,
  title={Marching cubes: A high resolution 3D surface construction algorithm},
  author={Lorensen, William E and Cline, Harvey E},
  booktitle={Seminal graphics: pioneering efforts that shaped the field},
  pages={347--353},
  year={1998}
}

@inproceedings{li2023sweetdreamer,
  author       = {Weiyu Li and
                  Rui Chen and
                  Xuelin Chen and
                  Ping Tan},
  title        = {SweetDreamer: Aligning Geometric Priors in 2D diffusion for Consistent
                  Text-to-3D},
  booktitle    = {The Twelfth International Conference on Learning Representations,
                  {ICLR} 2024, Vienna, Austria, May 7-11, 2024},
  publisher    = {OpenReview.net},
  year         = {2024},
  url          = {https://openreview.net/forum?id=extpNXo6hB},
  bibsource    = {dblp computer science bibliography, https://dblp.org}
}

@article{deitke2023objaverse-xl,
  title={Objaverse-xl: A universe of 10m+ 3d objects},
  author={Deitke, Matt and Liu, Ruoshi and Wallingford, Matthew and Ngo, Huong and Michel, Oscar and Kusupati, Aditya and Fan, Alan and Laforte, Christian and Voleti, Vikram and Gadre, Samir Yitzhak and others},
  journal={Advances in Neural Information Processing Systems},
  volume={36},
  pages={35799--35813},
  year={2023}
}

@article{3d-future,
  author       = {Huan Fu and
                  Rongfei Jia and
                  Lin Gao and
                  Mingming Gong and
                  Binqiang Zhao and
                  Stephen J. Maybank and
                  Dacheng Tao},
  title        = {3D-FUTURE: 3D Furniture Shape with TextURE},
  journal      = {Int. J. Comput. Vis.},
  volume       = {129},
  number       = {12},
  pages        = {3313--3337},
  year         = {2021},
  url          = {https://doi.org/10.1007/s11263-021-01534-z},
  doi          = {10.1007/S11263-021-01534-Z},
  bibsource    = {dblp computer science bibliography, https://dblp.org}
}

@article{3dshape2vecset,
  author       = {Biao Zhang and
                  Jiapeng Tang and
                  Matthias Nie{\ss}ner and
                  Peter Wonka},
  title        = {3DShape2VecSet: {A} 3D Shape Representation for Neural Fields and
                  Generative Diffusion Models},
  journal      = {{ACM} Trans. Graph.},
  volume       = {42},
  number       = {4},
  pages        = {92:1--92:16},
  year         = {2023},
  url          = {https://doi.org/10.1145/3592442},
  doi          = {10.1145/3592442},
  bibsource    = {dblp computer science bibliography, https://dblp.org}
}

@article{clay,
  author       = {Longwen Zhang and
                  Ziyu Wang and
                  Qixuan Zhang and
                  Qiwei Qiu and
                  Anqi Pang and
                  Haoran Jiang and
                  Wei Yang and
                  Lan Xu and
                  Jingyi Yu},
  title        = {{CLAY:} {A} Controllable Large-scale Generative Model for Creating
                  High-quality 3D Assets},
  journal      = {{ACM} Trans. Graph.},
  volume       = {43},
  number       = {4},
  pages        = {120:1--120:20},
  year         = {2024},
  url          = {https://doi.org/10.1145/3658146},
  doi          = {10.1145/3658146},
  bibsource    = {dblp computer science bibliography, https://dblp.org}
}

@article{triposg,
  title={Triposg: High-fidelity 3d shape synthesis using large-scale rectified flow models},
  author={Li, Yangguang and Zou, Zi-Xin and Liu, Zexiang and Wang, Dehu and Liang, Yuan and Yu, Zhipeng and Liu, Xingchao and Guo, Yuan-Chen and Liang, Ding and Ouyang, Wanli and others},
  journal={arXiv preprint arXiv:2502.06608},
  year={2025}
}

@article{hunyuan3d,
  title={Hunyuan3d 2.0: Scaling diffusion models for high resolution textured 3d assets generation},
  author={Zhao, Zibo and Lai, Zeqiang and Lin, Qingxiang and Zhao, Yunfei and Liu, Haolin and Yang, Shuhui and Feng, Yifei and Yang, Mingxin and Zhang, Sheng and Yang, Xianghui and others},
  journal={arXiv preprint arXiv:2501.12202},
  year={2025}
}

@inproceedings{trellis,
  author       = {Jianfeng Xiang and
                  Zelong Lv and
                  Sicheng Xu and
                  Yu Deng and
                  Ruicheng Wang and
                  Bowen Zhang and
                  Dong Chen and
                  Xin Tong and
                  Jiaolong Yang},
  title        = {Structured 3D Latents for Scalable and Versatile 3D Generation},
  booktitle    = {{IEEE/CVF} Conference on Computer Vision and Pattern Recognition,
                  {CVPR} 2025, Nashville, TN, USA, June 11-15, 2025},
  pages        = {21469--21480},
  publisher    = {Computer Vision Foundation / {IEEE}},
  year         = {2025},
  url          = {https://openaccess.thecvf.com/content/CVPR2025/html/Xiang\_Structured\_3D\_Latents\_for\_Scalable\_and\_Versatile\_3D\_Generation\_CVPR\_2025\_paper.html},
  doi          = {10.1109/CVPR52734.2025.02000},
  bibsource    = {dblp computer science bibliography, https://dblp.org}
}

@inproceedings{xcube,
  author       = {Xuanchi Ren and
                  Jiahui Huang and
                  Xiaohui Zeng and
                  Ken Museth and
                  Sanja Fidler and
                  Francis Williams},
  title        = {XCube: Large-Scale 3D Generative Modeling using Sparse Voxel Hierarchies},
  booktitle    = {{IEEE/CVF} Conference on Computer Vision and Pattern Recognition,
                  {CVPR} 2024, Seattle, WA, USA, June 16-22, 2024},
  pages        = {4209--4219},
  publisher    = {{IEEE}},
  year         = {2024},
  url          = {https://doi.org/10.1109/CVPR52733.2024.00403},
  doi          = {10.1109/CVPR52733.2024.00403},
  bibsource    = {dblp computer science bibliography, https://dblp.org}
}

@article{direct3d-s2,
  title={Direct3d-s2: Gigascale 3d generation made easy with spatial sparse attention},
  author={Wu, Shuang and Lin, Youtian and Zhang, Feihu and Zeng, Yifei and Yang, Yikang and Bao, Yajie and Qian, Jiachen and Zhu, Siyu and Cao, Xun and Torr, Philip and others},
  journal={arXiv preprint arXiv:2505.17412},
  year={2025}
}

@article{sparc3d,
  title={Sparc3D: Sparse Representation and Construction for High-Resolution 3D Shapes Modeling},
  author={Li, Zhihao and Wang, Yufei and Zheng, Heliang and Luo, Yihao and Wen, Bihan},
  journal={arXiv preprint arXiv:2505.14521},
  year={2025}
}

@inproceedings{wang2023prolificdreamerhighfidelitydiversetextto3d,
  author       = {Zhengyi Wang and
                  Cheng Lu and
                  Yikai Wang and
                  Fan Bao and
                  Chongxuan Li and
                  Hang Su and
                  Jun Zhu},
  editor       = {Alice Oh and
                  Tristan Naumann and
                  Amir Globerson and
                  Kate Saenko and
                  Moritz Hardt and
                  Sergey Levine},
  title        = {ProlificDreamer: High-Fidelity and Diverse Text-to-3D Generation with
                  Variational Score Distillation},
  booktitle    = {Advances in Neural Information Processing Systems 36: Annual Conference
                  on Neural Information Processing Systems 2023, NeurIPS 2023, New Orleans,
                  LA, USA, December 10 - 16, 2023},
  year         = {2023},
  url          = {http://papers.nips.cc/paper\_files/paper/2023/hash/1a87980b9853e84dfb295855b425c262-Abstract-Conference.html},
  bibsource    = {dblp computer science bibliography, https://dblp.org}
}

@inproceedings{liu2023zero1to3zeroshotimage3d,
  author       = {Ruoshi Liu and
                  Rundi Wu and
                  Basile Van Hoorick and
                  Pavel Tokmakov and
                  Sergey Zakharov and
                  Carl Vondrick},
  title        = {Zero-1-to-3: Zero-shot One Image to 3D Object},
  booktitle    = {{IEEE/CVF} International Conference on Computer Vision, {ICCV} 2023,
                  Paris, France, October 1-6, 2023},
  pages        = {9264--9275},
  publisher    = {{IEEE}},
  year         = {2023},
  url          = {https://doi.org/10.1109/ICCV51070.2023.00853},
  doi          = {10.1109/ICCV51070.2023.00853},
  bibsource    = {dblp computer science bibliography, https://dblp.org}
}

@inproceedings{shi2024mvdreammultiviewdiffusion3d,
  author       = {Yichun Shi and
                  Peng Wang and
                  Jianglong Ye and
                  Long Mai and
                  Kejie Li and
                  Xiao Yang},
  title        = {MVDream: Multi-view Diffusion for 3D Generation},
  booktitle    = {The Twelfth International Conference on Learning Representations,
                  {ICLR} 2024, Vienna, Austria, May 7-11, 2024},
  publisher    = {OpenReview.net},
  year         = {2024},
  url          = {https://openreview.net/forum?id=FUgrjq2pbB},
  bibsource    = {dblp computer science bibliography, https://dblp.org}
}

@article{chang20243d,
  title={3D Shape Tokenization via Latent Flow Matching},
  author={Chang, Jen-Hao Rick and Wang, Yuyang and Martin, Miguel Angel Bautista and Gu, Jiatao and Zhao, Xiaoming and Susskind, Josh and Tuzel, Oncel},
  journal={arXiv preprint arXiv:2412.15618},
  year={2024}
}

@inproceedings{gao2022get3dgenerativemodelhigh,
  author       = {Jun Gao and
                  Tianchang Shen and
                  Zian Wang and
                  Wenzheng Chen and
                  Kangxue Yin and
                  Daiqing Li and
                  Or Litany and
                  Zan Gojcic and
                  Sanja Fidler},
  editor       = {Sanmi Koyejo and
                  S. Mohamed and
                  A. Agarwal and
                  Danielle Belgrave and
                  K. Cho and
                  A. Oh},
  title        = {{GET3D:} {A} Generative Model of High Quality 3D Textured Shapes Learned
                  from Images},
  booktitle    = {Advances in Neural Information Processing Systems 35: Annual Conference
                  on Neural Information Processing Systems 2022, NeurIPS 2022, New Orleans,
                  LA, USA, November 28 - December 9, 2022},
  year         = {2022},
  url          = {http://papers.nips.cc/paper\_files/paper/2022/hash/cebbd24f1e50bcb63d015611fe0fe767-Abstract-Conference.html},
  bibsource    = {dblp computer science bibliography, https://dblp.org}
}

@article{jun2023shapegeneratingconditional3d,
  title={Shap-e: Generating conditional 3d implicit functions},
  author={Jun, Heewoo and Nichol, Alex},
  journal={arXiv preprint arXiv:2305.02463},
  year={2023}
}

@article{nichol2022pointegenerating3dpoint,
  title={Point-e: A system for generating 3d point clouds from complex prompts},
  author={Nichol, Alex and Jun, Heewoo and Dhariwal, Prafulla and Mishkin, Pamela and Chen, Mark},
  journal={arXiv preprint arXiv:2212.08751},
  year={2022}
}

@article{ye2025hi3dgen,
  title={Hi3dgen: High-fidelity 3d geometry generation from images via normal bridging},
  author={Ye, Chongjie and Wu, Yushuang and Lu, Ziteng and Chang, Jiahao and Guo, Xiaoyang and Zhou, Jiaqing and Zhao, Hao and Han, Xiaoguang}
}

@article{liu2025quadgptnativequadrilateralmesh,
  title={QuadGPT: Native Quadrilateral Mesh Generation with Autoregressive Models},
  author={Liu, Jian and Wang, Chunshi and Guo, Song and Weng, Haohan and Zhou, Zhen and Li, Zhiqi and Yu, Jiaao and Zhu, Yiling and Xu, Jing and Lei, Biwen and others},
  journal={arXiv preprint arXiv:2509.21420},
  year={2025}
}

@article{liu2025meshrftenhancingmeshgeneration,
  title={Mesh-RFT: Enhancing Mesh Generation via Fine-grained Reinforcement Fine-Tuning},
  author={Liu, Jian and Xu, Jing and Guo, Song and Li, Jing and Guo, Jingfeng and Yu, Jiaao and Weng, Haohan and Lei, Biwen and Yang, Xianghui and Chen, Zhuo and others},
  journal={arXiv preprint arXiv:2505.16761},
  year={2025}
}

@inproceedings{yuan2025nativesparseattention,
  author       = {Jingyang Yuan and
                  Huazuo Gao and
                  Damai Dai and
                  Junyu Luo and
                  Liang Zhao and
                  Zhengyan Zhang and
                  Zhenda Xie and
                  Yuxing Wei and
                  Lean Wang and
                  Zhiping Xiao and
                  Yuqing Wang and
                  Chong Ruan and
                  Ming Zhang and
                  Wenfeng Liang and
                  Wangding Zeng},
  editor       = {Wanxiang Che and
                  Joyce Nabende and
                  Ekaterina Shutova and
                  Mohammad Taher Pilehvar},
  title        = {Native Sparse Attention: Hardware-Aligned and Natively Trainable Sparse
                  Attention},
  booktitle    = {Proceedings of the 63rd Annual Meeting of the Association for Computational
                  Linguistics (Volume 1: Long Papers), {ACL} 2025, Vienna, Austria,
                  July 27 - August 1, 2025},
  pages        = {23078--23097},
  publisher    = {Association for Computational Linguistics},
  year         = {2025},
  url          = {https://aclanthology.org/2025.acl-long.1126/},
  bibsource    = {dblp computer science bibliography, https://dblp.org}
}

@inproceedings{huan20253dfront,
  author       = {Huan Fu and
                  Bowen Cai and
                  Lin Gao and
                  Lingxiao Zhang and
                  Jiaming Wang and
                  Cao Li and
                  Qixun Zeng and
                  Chengyue Sun and
                  Rongfei Jia and
                  Binqiang Zhao and
                  Hao Zhang},
  title        = {3D-FRONT: 3D Furnished Rooms with layOuts and semaNTics},
  booktitle    = {2021 {IEEE/CVF} International Conference on Computer Vision, {ICCV}
                  2021, Montreal, QC, Canada, October 10-17, 2021},
  pages        = {10913--10922},
  publisher    = {{IEEE}},
  year         = {2021},
  url          = {https://doi.org/10.1109/ICCV48922.2021.01075},
  doi          = {10.1109/ICCV48922.2021.01075},
  bibsource    = {dblp computer science bibliography, https://dblp.org}
}

@inproceedings{DBLP:Fantasia3D,
  author       = {Rui Chen and
                  Yongwei Chen and
                  Ningxin Jiao and
                  Kui Jia},
  title        = {Fantasia3D: Disentangling Geometry and Appearance for High-quality
                  Text-to-3D Content Creation},
  booktitle    = {{IEEE/CVF} International Conference on Computer Vision, {ICCV} 2023,
                  Paris, France, October 1-6, 2023},
  pages        = {22189--22199},
  publisher    = {{IEEE}},
  year         = {2023},
  url          = {https://doi.org/10.1109/ICCV51070.2023.02033},
  doi          = {10.1109/ICCV51070.2023.02033},
  bibsource    = {dblp computer science bibliography, https://dblp.org}
}

@inproceedings{DBLP:Text-to-3D,
  author       = {Zilong Chen and
                  Feng Wang and
                  Yikai Wang and
                  Huaping Liu},
  title        = {Text-to-3D using Gaussian Splatting},
  booktitle    = {{IEEE/CVF} Conference on Computer Vision and Pattern Recognition,
                  {CVPR} 2024, Seattle, WA, USA, June 16-22, 2024},
  pages        = {21401--21412},
  publisher    = {{IEEE}},
  year         = {2024},
  url          = {https://doi.org/10.1109/CVPR52733.2024.02022},
  doi          = {10.1109/CVPR52733.2024.02022},
  bibsource    = {dblp computer science bibliography, https://dblp.org}
}

@inproceedings{DBLP:Magic3D,
  author       = {Chen{-}Hsuan Lin and
                  Jun Gao and
                  Luming Tang and
                  Towaki Takikawa and
                  Xiaohui Zeng and
                  Xun Huang and
                  Karsten Kreis and
                  Sanja Fidler and
                  Ming{-}Yu Liu and
                  Tsung{-}Yi Lin},
  title        = {Magic3D: High-Resolution Text-to-3D Content Creation},
  booktitle    = {{IEEE/CVF} Conference on Computer Vision and Pattern Recognition,
                  {CVPR} 2023, Vancouver, BC, Canada, June 17-24, 2023},
  pages        = {300--309},
  publisher    = {{IEEE}},
  year         = {2023},
  url          = {https://doi.org/10.1109/CVPR52729.2023.00037},
  doi          = {10.1109/CVPR52729.2023.00037},
  bibsource    = {dblp computer science bibliography, https://dblp.org}
}

@inproceedings{DBLP:ScaleDreamer,
  author       = {Zhiyuan Ma and
                  Yuxiang Wei and
                  Yabin Zhang and
                  Xiangyu Zhu and
                  Zhen Lei and
                  Lei Zhang},
  editor       = {Ales Leonardis and
                  Elisa Ricci and
                  Stefan Roth and
                  Olga Russakovsky and
                  Torsten Sattler and
                  G{\"{u}}l Varol},
  title        = {ScaleDreamer: Scalable Text-to-3D Synthesis with Asynchronous Score
                  Distillation},
  booktitle    = {Computer Vision - {ECCV} 2024 - 18th European Conference, Milan, Italy,
                  September 29-October 4, 2024, Proceedings, Part {VII}},
  series       = {Lecture Notes in Computer Science},
  volume       = {15065},
  pages        = {1--19},
  publisher    = {Springer},
  year         = {2024},
  url          = {https://doi.org/10.1007/978-3-031-72667-5\_1},
  doi          = {10.1007/978-3-031-72667-5\_1},
  bibsource    = {dblp computer science bibliography, https://dblp.org}
}

@inproceedings{DBLP:DreamBooth3D,
  author       = {Amit Raj and
                  Srinivas Kaza and
                  Ben Poole and
                  Michael Niemeyer and
                  Nataniel Ruiz and
                  Ben Mildenhall and
                  Shiran Zada and
                  Kfir Aberman and
                  Michael Rubinstein and
                  Jonathan T. Barron and
                  Yuanzhen Li and
                  Varun Jampani},
  title        = {DreamBooth3D: Subject-Driven Text-to-3D Generation},
  booktitle    = {{IEEE/CVF} International Conference on Computer Vision, {ICCV} 2023,
                  Paris, France, October 1-6, 2023},
  pages        = {2349--2359},
  publisher    = {{IEEE}},
  year         = {2023},
  url          = {https://doi.org/10.1109/ICCV51070.2023.00223},
  doi          = {10.1109/ICCV51070.2023.00223},
  bibsource    = {dblp computer science bibliography, https://dblp.org}
}

@inproceedings{DBLP:DreamCraft3D,
  author       = {Jingxiang Sun and
                  Bo Zhang and
                  Ruizhi Shao and
                  Lizhen Wang and
                  Wen Liu and
                  Zhenda Xie and
                  Yebin Liu},
  title        = {DreamCraft3D: Hierarchical 3D Generation with Bootstrapped Diffusion
                  Prior},
  booktitle    = {The Twelfth International Conference on Learning Representations,
                  {ICLR} 2024, Vienna, Austria, May 7-11, 2024},
  publisher    = {OpenReview.net},
  year         = {2024},
  url          = {https://openreview.net/forum?id=DDX1u29Gqr},
  bibsource    = {dblp computer science bibliography, https://dblp.org}
}

@inproceedings{DBLP:DreamGaussian,
  author       = {Jiaxiang Tang and
                  Jiawei Ren and
                  Hang Zhou and
                  Ziwei Liu and
                  Gang Zeng},
  title        = {DreamGaussian: Generative Gaussian Splatting for Efficient 3D Content
                  Creation},
  booktitle    = {The Twelfth International Conference on Learning Representations,
                  {ICLR} 2024, Vienna, Austria, May 7-11, 2024},
  publisher    = {OpenReview.net},
  year         = {2024},
  url          = {https://openreview.net/forum?id=UyNXMqnN3c},
  bibsource    = {dblp computer science bibliography, https://dblp.org}
}

@inproceedings{DBLP:ScoreJacobianChaining,
  author       = {Haochen Wang and
                  Xiaodan Du and
                  Jiahao Li and
                  Raymond A. Yeh and
                  Greg Shakhnarovich},
  title        = {Score Jacobian Chaining: Lifting Pretrained 2D Diffusion Models for
                  3D Generation},
  booktitle    = {{IEEE/CVF} Conference on Computer Vision and Pattern Recognition,
                  {CVPR} 2023, Vancouver, BC, Canada, June 17-24, 2023},
  pages        = {12619--12629},
  publisher    = {{IEEE}},
  year         = {2023},
  url          = {https://doi.org/10.1109/CVPR52729.2023.01214},
  doi          = {10.1109/CVPR52729.2023.01214},
  bibsource    = {dblp computer science bibliography, https://dblp.org}
}

@inproceedings{DBLP:AnimatableDreamer,
  author       = {Xinzhou Wang and
                  Yikai Wang and
                  Junliang Ye and
                  Fuchun Sun and
                  Zhengyi Wang and
                  Ling Wang and
                  Pengkun Liu and
                  Kai Sun and
                  Xintong Wang and
                  Wende Xie and
                  Fangfu Liu and
                  Bin He},
  editor       = {Ales Leonardis and
                  Elisa Ricci and
                  Stefan Roth and
                  Olga Russakovsky and
                  Torsten Sattler and
                  G{\"{u}}l Varol},
  title        = {AnimatableDreamer: Text-Guided Non-rigid 3D Model Generation and Reconstruction
                  with Canonical Score Distillation},
  booktitle    = {Computer Vision - {ECCV} 2024 - 18th European Conference, Milan, Italy,
                  September 29-October 4, 2024, Proceedings, Part {XXV}},
  series       = {Lecture Notes in Computer Science},
  volume       = {15083},
  pages        = {321--339},
  publisher    = {Springer},
  year         = {2024},
  url          = {https://doi.org/10.1007/978-3-031-72698-9\_19},
  doi          = {10.1007/978-3-031-72698-9\_19},
  bibsource    = {dblp computer science bibliography, https://dblp.org}
}

@inproceedings{DBLP:GaussianDreamer,
  author       = {Taoran Yi and
                  Jiemin Fang and
                  Junjie Wang and
                  Guanjun Wu and
                  Lingxi Xie and
                  Xiaopeng Zhang and
                  Wenyu Liu and
                  Qi Tian and
                  Xinggang Wang},
  title        = {GaussianDreamer: Fast Generation from Text to 3D Gaussians by Bridging
                  2D and 3D Diffusion Models},
  booktitle    = {{IEEE/CVF} Conference on Computer Vision and Pattern Recognition,
                  {CVPR} 2024, Seattle, WA, USA, June 16-22, 2024},
  pages        = {6796--6807},
  publisher    = {{IEEE}},
  year         = {2024},
  url          = {https://doi.org/10.1109/CVPR52733.2024.00649},
  doi          = {10.1109/CVPR52733.2024.00649},
  bibsource    = {dblp computer science bibliography, https://dblp.org}
}

@article{wang2023imagedream,
  title={Imagedream: Image-prompt multi-view diffusion for 3d generation},
  author={Wang, Peng and Shi, Yichun},
  journal={arXiv preprint arXiv:2312.02201},
  year={2023}
}

@inproceedings{DBLP:DreamReward,
  author       = {Junliang Ye and
                  Fangfu Liu and
                  Qixiu Li and
                  Zhengyi Wang and
                  Yikai Wang and
                  Xinzhou Wang and
                  Yueqi Duan and
                  Jun Zhu},
  editor       = {Ales Leonardis and
                  Elisa Ricci and
                  Stefan Roth and
                  Olga Russakovsky and
                  Torsten Sattler and
                  G{\"{u}}l Varol},
  title        = {DreamReward: Text-to-3D Generation with Human Preference},
  booktitle    = {Computer Vision - {ECCV} 2024 - 18th European Conference, Milan, Italy,
                  September 29-October 4, 2024, Proceedings, Part {LXX}},
  series       = {Lecture Notes in Computer Science},
  volume       = {15128},
  pages        = {259--276},
  publisher    = {Springer},
  year         = {2024},
  url          = {https://doi.org/10.1007/978-3-031-72897-6\_15},
  doi          = {10.1007/978-3-031-72897-6\_15},
  bibsource    = {dblp computer science bibliography, https://dblp.org}
}

@article{chen2024v3d,
  title={V3d: Video diffusion models are effective 3d generators},
  author={Chen, Zilong and Wang, Yikai and Wang, Feng and Wang, Zhengyi and Liu, Huaping},
  journal={arXiv preprint arXiv:2403.06738},
  year={2024}
}

@inproceedings{DBLP:Animate3D,
  author       = {Yanqin Jiang and
                  Chaohui Yu and
                  Chenjie Cao and
                  Fan Wang and
                  Weiming Hu and
                  Jin Gao},
  editor       = {Amir Globersons and
                  Lester Mackey and
                  Danielle Belgrave and
                  Angela Fan and
                  Ulrich Paquet and
                  Jakub M. Tomczak and
                  Cheng Zhang},
  title        = {Animate3D: Animating Any 3D Model with Multi-view Video Diffusion},
  booktitle    = {Advances in Neural Information Processing Systems 38: Annual Conference
                  on Neural Information Processing Systems 2024, NeurIPS 2024, Vancouver,
                  BC, Canada, December 10 - 15, 2024},
  year         = {2024},
  url          = {http://papers.nips.cc/paper\_files/paper/2024/hash/e3b53f89136b1bc69a5714ea465f01b6-Abstract-Conference.html},
  bibsource    = {dblp computer science bibliography, https://dblp.org}
}

@inproceedings{DBLP:One-2-3-45,
  author       = {Minghua Liu and
                  Chao Xu and
                  Haian Jin and
                  Linghao Chen and
                  Mukund Varma T. and
                  Zexiang Xu and
                  Hao Su},
  editor       = {Alice Oh and
                  Tristan Naumann and
                  Amir Globerson and
                  Kate Saenko and
                  Moritz Hardt and
                  Sergey Levine},
  title        = {One-2-3-45: Any Single Image to 3D Mesh in 45 Seconds without Per-Shape
                  Optimization},
  booktitle    = {Advances in Neural Information Processing Systems 36: Annual Conference
                  on Neural Information Processing Systems 2023, NeurIPS 2023, New Orleans,
                  LA, USA, December 10 - 16, 2023},
  year         = {2023},
  url          = {http://papers.nips.cc/paper\_files/paper/2023/hash/4683beb6bab325650db13afd05d1a14a-Abstract-Conference.html},
  bibsource    = {dblp computer science bibliography, https://dblp.org}
}

@inproceedings{DBLP:SyncDreamer,
  author       = {Yuan Liu and
                  Cheng Lin and
                  Zijiao Zeng and
                  Xiaoxiao Long and
                  Lingjie Liu and
                  Taku Komura and
                  Wenping Wang},
  title        = {SyncDreamer: Generating Multiview-consistent Images from a Single-view
                  Image},
  booktitle    = {The Twelfth International Conference on Learning Representations,
                  {ICLR} 2024, Vienna, Austria, May 7-11, 2024},
  publisher    = {OpenReview.net},
  year         = {2024},
  url          = {https://openreview.net/forum?id=MN3yH2ovHb},
  bibsource    = {dblp computer science bibliography, https://dblp.org}
}

@inproceedings{DBLP:Wonder3D,
  author       = {Xiaoxiao Long and
                  Yuan{-}Chen Guo and
                  Cheng Lin and
                  Yuan Liu and
                  Zhiyang Dou and
                  Lingjie Liu and
                  Yuexin Ma and
                  Song{-}Hai Zhang and
                  Marc Habermann and
                  Christian Theobalt and
                  Wenping Wang},
  title        = {Wonder3D: Single Image to 3D Using Cross-Domain Diffusion},
  booktitle    = {{IEEE/CVF} Conference on Computer Vision and Pattern Recognition,
                  {CVPR} 2024, Seattle, WA, USA, June 16-22, 2024},
  pages        = {9970--9980},
  publisher    = {{IEEE}},
  year         = {2024},
  url          = {https://doi.org/10.1109/CVPR52733.2024.00951},
  doi          = {10.1109/CVPR52733.2024.00951},
  bibsource    = {dblp computer science bibliography, https://dblp.org}
}

@article{shi2023zero123++,
  title={Zero123++: a single image to consistent multi-view diffusion base model},
  author={Shi, Ruoxi and Chen, Hansheng and Zhang, Zhuoyang and Liu, Minghua and Xu, Chao and Wei, Xinyue and Chen, Linghao and Zeng, Chong and Su, Hao},
  journal={arXiv preprint arXiv:2310.15110},
  year={2023}
}

@inproceedings{DBLP:SV3D,
  author       = {Vikram Voleti and
                  Chun{-}Han Yao and
                  Mark Boss and
                  Adam Letts and
                  David Pankratz and
                  Dmitry Tochilkin and
                  Christian Laforte and
                  Robin Rombach and
                  Varun Jampani},
  editor       = {Ales Leonardis and
                  Elisa Ricci and
                  Stefan Roth and
                  Olga Russakovsky and
                  Torsten Sattler and
                  G{\"{u}}l Varol},
  title        = {{SV3D:} Novel Multi-view Synthesis and 3D Generation from a Single
                  Image Using Latent Video Diffusion},
  booktitle    = {Computer Vision - {ECCV} 2024 - 18th European Conference, Milan, Italy,
                  September 29-October 4, 2024, Proceedings, Part {I}},
  series       = {Lecture Notes in Computer Science},
  volume       = {15059},
  pages        = {439--457},
  publisher    = {Springer},
  year         = {2024},
  url          = {https://doi.org/10.1007/978-3-031-73232-4\_25},
  doi          = {10.1007/978-3-031-73232-4\_25},
  bibsource    = {dblp computer science bibliography, https://dblp.org}
}

@inproceedings{DBLP:Consistent123,
  author       = {Yukang Lin and
                  Haonan Han and
                  Chaoqun Gong and
                  Zunnan Xu and
                  Yachao Zhang and
                  Xiu Li},
  editor       = {Jianfei Cai and
                  Mohan S. Kankanhalli and
                  Balakrishnan Prabhakaran and
                  Susanne Boll and
                  Ramanathan Subramanian and
                  Liang Zheng and
                  Vivek K. Singh and
                  Pablo C{\'{e}}sar and
                  Lexing Xie and
                  Dong Xu},
  title        = {Consistent123: One Image to Highly Consistent 3D Asset Using Case-Aware
                  Diffusion Priors},
  booktitle    = {Proceedings of the 32nd {ACM} International Conference on Multimedia,
                  {MM} 2024, Melbourne, VIC, Australia, 28 October 2024 - 1 November
                  2024},
  pages        = {6715--6724},
  publisher    = {{ACM}},
  year         = {2024},
  url          = {https://doi.org/10.1145/3664647.3680994},
  doi          = {10.1145/3664647.3680994},
  bibsource    = {dblp computer science bibliography, https://dblp.org}
}

@inproceedings{DBLP:Unique3D,
  author       = {Kailu Wu and
                  Fangfu Liu and
                  Zhihan Cai and
                  Runjie Yan and
                  Hanyang Wang and
                  Yating Hu and
                  Yueqi Duan and
                  Kaisheng Ma},
  editor       = {Amir Globersons and
                  Lester Mackey and
                  Danielle Belgrave and
                  Angela Fan and
                  Ulrich Paquet and
                  Jakub M. Tomczak and
                  Cheng Zhang},
  title        = {Unique3D: High-Quality and Efficient 3D Mesh Generation from a Single
                  Image},
  booktitle    = {Advances in Neural Information Processing Systems 38: Annual Conference
                  on Neural Information Processing Systems 2024, NeurIPS 2024, Vancouver,
                  BC, Canada, December 10 - 15, 2024},
  year         = {2024},
  url          = {http://papers.nips.cc/paper\_files/paper/2024/hash/e25198b6a75f74277ee3a2bd4165d9ef-Abstract-Conference.html},
  bibsource    = {dblp computer science bibliography, https://dblp.org}
}

@inproceedings{DBLP:Hi3D,
  author       = {Haibo Yang and
                  Yang Chen and
                  Yingwei Pan and
                  Ting Yao and
                  Zhineng Chen and
                  Chong{-}Wah Ngo and
                  Tao Mei},
  editor       = {Jianfei Cai and
                  Mohan S. Kankanhalli and
                  Balakrishnan Prabhakaran and
                  Susanne Boll and
                  Ramanathan Subramanian and
                  Liang Zheng and
                  Vivek K. Singh and
                  Pablo C{\'{e}}sar and
                  Lexing Xie and
                  Dong Xu},
  title        = {Hi3D: Pursuing High-Resolution Image-to-3D Generation with Video Diffusion
                  Models},
  booktitle    = {Proceedings of the 32nd {ACM} International Conference on Multimedia,
                  {MM} 2024, Melbourne, VIC, Australia, 28 October 2024 - 1 November
                  2024},
  pages        = {6870--6879},
  publisher    = {{ACM}},
  year         = {2024},
  url          = {https://doi.org/10.1145/3664647.3681634},
  doi          = {10.1145/3664647.3681634},
  bibsource    = {dblp computer science bibliography, https://dblp.org}
}

@article{zhao2024flexidreamer,
  title={Flexidreamer: single image-to-3d generation with flexicubes},
  author={Zhao, Ruowen and Wang, Zhengyi and Wang, Yikai and Zhou, Zihan and Zhu, Jun},
  journal={arXiv preprint arXiv:2404.00987},
  year={2024}
}

@article{DBLP:NeRF,
  author       = {Ben Mildenhall and
                  Pratul P. Srinivasan and
                  Matthew Tancik and
                  Jonathan T. Barron and
                  Ravi Ramamoorthi and
                  Ren Ng},
  title        = {NeRF: representing scenes as neural radiance fields for view synthesis},
  journal      = {Commun. {ACM}},
  volume       = {65},
  number       = {1},
  pages        = {99--106},
  year         = {2022},
  url          = {https://doi.org/10.1145/3503250},
  doi          = {10.1145/3503250},
  bibsource    = {dblp computer science bibliography, https://dblp.org}
}

@inproceedings{DBLP:Meta3DAssetGen,
  author       = {Yawar Siddiqui and
                  Tom Monnier and
                  Filippos Kokkinos and
                  Mahendra Kariya and
                  Yanir Kleiman and
                  Emilien Garreau and
                  Oran Gafni and
                  Natalia Neverova and
                  Andrea Vedaldi and
                  Roman Shapovalov and
                  David Novotn{\'{y}}},
  editor       = {Amir Globersons and
                  Lester Mackey and
                  Danielle Belgrave and
                  Angela Fan and
                  Ulrich Paquet and
                  Jakub M. Tomczak and
                  Cheng Zhang},
  title        = {Meta 3D AssetGen: Text-to-Mesh Generation with High-Quality Geometry,
                  Texture, and {PBR} Materials},
  booktitle    = {Advances in Neural Information Processing Systems 38: Annual Conference
                  on Neural Information Processing Systems 2024, NeurIPS 2024, Vancouver,
                  BC, Canada, December 10 - 15, 2024},
  year         = {2024},
  url          = {http://papers.nips.cc/paper\_files/paper/2024/hash/123cfe7d8b7702ac97aaf4468fc05fa5-Abstract-Conference.html},
  bibsource    = {dblp computer science bibliography, https://dblp.org}
}

@inproceedings{DBLP:LGM,
  author       = {Jiaxiang Tang and
                  Zhaoxi Chen and
                  Xiaokang Chen and
                  Tengfei Wang and
                  Gang Zeng and
                  Ziwei Liu},
  editor       = {Ales Leonardis and
                  Elisa Ricci and
                  Stefan Roth and
                  Olga Russakovsky and
                  Torsten Sattler and
                  G{\"{u}}l Varol},
  title        = {{LGM:} Large Multi-view Gaussian Model for High-Resolution 3D Content
                  Creation},
  booktitle    = {Computer Vision - {ECCV} 2024 - 18th European Conference, Milan, Italy,
                  September 29-October 4, 2024, Proceedings, Part {IV}},
  series       = {Lecture Notes in Computer Science},
  volume       = {15062},
  pages        = {1--18},
  publisher    = {Springer},
  year         = {2024},
  url          = {https://doi.org/10.1007/978-3-031-73235-5\_1},
  doi          = {10.1007/978-3-031-73235-5\_1},
  bibsource    = {dblp computer science bibliography, https://dblp.org}
}

@inproceedings{DBLP:PF-LRM,
  author       = {Peng Wang and
                  Hao Tan and
                  Sai Bi and
                  Yinghao Xu and
                  Fujun Luan and
                  Kalyan Sunkavalli and
                  Wenping Wang and
                  Zexiang Xu and
                  Kai Zhang},
  title        = {{PF-LRM:} Pose-Free Large Reconstruction Model for Joint Pose and
                  Shape Prediction},
  booktitle    = {The Twelfth International Conference on Learning Representations,
                  {ICLR} 2024, Vienna, Austria, May 7-11, 2024},
  publisher    = {OpenReview.net},
  year         = {2024},
  url          = {https://openreview.net/forum?id=noe76eRcPC},
  bibsource    = {dblp computer science bibliography, https://dblp.org}
}

@inproceedings{DBLP:CRM,
  author       = {Zhengyi Wang and
                  Yikai Wang and
                  Yifei Chen and
                  Chendong Xiang and
                  Shuo Chen and
                  Dajiang Yu and
                  Chongxuan Li and
                  Hang Su and
                  Jun Zhu},
  editor       = {Ales Leonardis and
                  Elisa Ricci and
                  Stefan Roth and
                  Olga Russakovsky and
                  Torsten Sattler and
                  G{\"{u}}l Varol},
  title        = {{CRM:} Single Image to 3D Textured Mesh with Convolutional Reconstruction
                  Model},
  booktitle    = {Computer Vision - {ECCV} 2024 - 18th European Conference, Milan, Italy,
                  September 29-October 4, 2024, Proceedings, Part {XXXI}},
  series       = {Lecture Notes in Computer Science},
  volume       = {15089},
  pages        = {57--74},
  publisher    = {Springer},
  year         = {2024},
  url          = {https://doi.org/10.1007/978-3-031-72751-1\_4},
  doi          = {10.1007/978-3-031-72751-1\_4},
  bibsource    = {dblp computer science bibliography, https://dblp.org}
}

@inproceedings{DBLP:DMV3D,
  author       = {Yinghao Xu and
                  Hao Tan and
                  Fujun Luan and
                  Sai Bi and
                  Peng Wang and
                  Jiahao Li and
                  Zifan Shi and
                  Kalyan Sunkavalli and
                  Gordon Wetzstein and
                  Zexiang Xu and
                  Kai Zhang},
  title        = {{DMV3D:} Denoising Multi-view Diffusion Using 3D Large Reconstruction
                  Model},
  booktitle    = {The Twelfth International Conference on Learning Representations,
                  {ICLR} 2024, Vienna, Austria, May 7-11, 2024},
  publisher    = {OpenReview.net},
  year         = {2024},
  url          = {https://openreview.net/forum?id=H4yQefeXhp},
  bibsource    = {dblp computer science bibliography, https://dblp.org}
}

@inproceedings{DBLP:GRM,
  author       = {Yinghao Xu and
                  Zifan Shi and
                  Yifan Wang and
                  Hansheng Chen and
                  Ceyuan Yang and
                  Sida Peng and
                  Yujun Shen and
                  Gordon Wetzstein},
  editor       = {Ales Leonardis and
                  Elisa Ricci and
                  Stefan Roth and
                  Olga Russakovsky and
                  Torsten Sattler and
                  G{\"{u}}l Varol},
  title        = {{GRM:} Large Gaussian Reconstruction Model for Efficient 3D Reconstruction
                  and Generation},
  booktitle    = {Computer Vision - {ECCV} 2024 - 18th European Conference, Milan, Italy,
                  September 29-October 4, 2024, Proceedings, Part {XV}},
  series       = {Lecture Notes in Computer Science},
  volume       = {15073},
  pages        = {1--20},
  publisher    = {Springer},
  year         = {2024},
  url          = {https://doi.org/10.1007/978-3-031-72633-0\_1},
  doi          = {10.1007/978-3-031-72633-0\_1},
  bibsource    = {dblp computer science bibliography, https://dblp.org}
}

@inproceedings{DBLP:GeoLRM,
  author       = {Chubin Zhang and
                  Hongliang Song and
                  Yi Wei and
                  Chen Yu and
                  Jiwen Lu and
                  Yansong Tang},
  editor       = {Amir Globersons and
                  Lester Mackey and
                  Danielle Belgrave and
                  Angela Fan and
                  Ulrich Paquet and
                  Jakub M. Tomczak and
                  Cheng Zhang},
  title        = {GeoLRM: Geometry-Aware Large Reconstruction Model for High-Quality
                  3D Gaussian Generation},
  booktitle    = {Advances in Neural Information Processing Systems 38: Annual Conference
                  on Neural Information Processing Systems 2024, NeurIPS 2024, Vancouver,
                  BC, Canada, December 10 - 15, 2024},
  year         = {2024},
  url          = {http://papers.nips.cc/paper\_files/paper/2024/hash/6506964d22ede4d36adae956e6a9919a-Abstract-Conference.html},
  bibsource    = {dblp computer science bibliography, https://dblp.org}
}

@inproceedings{DBLP:GS-LRM,
  author       = {Kai Zhang and
                  Sai Bi and
                  Hao Tan and
                  Yuanbo Xiangli and
                  Nanxuan Zhao and
                  Kalyan Sunkavalli and
                  Zexiang Xu},
  editor       = {Ales Leonardis and
                  Elisa Ricci and
                  Stefan Roth and
                  Olga Russakovsky and
                  Torsten Sattler and
                  G{\"{u}}l Varol},
  title        = {{GS-LRM:} Large Reconstruction Model for 3D Gaussian Splatting},
  booktitle    = {Computer Vision - {ECCV} 2024 - 18th European Conference, Milan, Italy,
                  September 29-October 4, 2024, Proceedings, Part {XXII}},
  series       = {Lecture Notes in Computer Science},
  volume       = {15080},
  pages        = {1--19},
  publisher    = {Springer},
  year         = {2024},
  url          = {https://doi.org/10.1007/978-3-031-72670-5\_1},
  doi          = {10.1007/978-3-031-72670-5\_1},
  bibsource    = {dblp computer science bibliography, https://dblp.org}
}

@inproceedings{ziwen2025longlrm,
  title={Long-lrm: Long-sequence large reconstruction model for wide-coverage gaussian splats},
  author={Ziwen, Chen and Tan, Hao and Zhang, Kai and Bi, Sai and Luan, Fujun and Hong, Yicong and Fuxin, Li and Xu, Zexiang},
  booktitle={Proceedings of the IEEE/CVF International Conference on Computer Vision},
  pages={4349--4359},
  year={2025}
}

@inproceedings{DBLP:TriplaneMeetsGaussianSplatting,
  author       = {Zi{-}Xin Zou and
                  Zhipeng Yu and
                  Yuan{-}Chen Guo and
                  Yangguang Li and
                  Ding Liang and
                  Yan{-}Pei Cao and
                  Song{-}Hai Zhang},
  title        = {Triplane Meets Gaussian Splatting: Fast and Generalizable Single-View
                  3D Reconstruction with Transformers},
  booktitle    = {{IEEE/CVF} Conference on Computer Vision and Pattern Recognition,
                  {CVPR} 2024, Seattle, WA, USA, June 16-22, 2024},
  pages        = {10324--10335},
  publisher    = {{IEEE}},
  year         = {2024},
  url          = {https://doi.org/10.1109/CVPR52733.2024.00983},
  doi          = {10.1109/CVPR52733.2024.00983},
  bibsource    = {dblp computer science bibliography, https://dblp.org}
}

@article{li2024craftsman3d,
  title={Craftsman3d: High-fidelity mesh generation with 3d native generation and interactive geometry refiner},
  author={Li, Weiyu and Liu, Jiarui and Yan, Hongyu and Chen, Rui and Liang, Yixun and Chen, Xuelin and Tan, Ping and Long, Xiaoxiao},
  journal={arXiv preprint arXiv:2405.14979},
  year={2024}
}
